\documentclass{article}
\usepackage{arxiv}
\pdfoutput=1

\usepackage[utf8]{inputenc} 
\usepackage[T1]{fontenc}    
\usepackage{hyperref}       
\usepackage{url}            
\usepackage{booktabs}       
\usepackage{amsfonts}       
\usepackage{nicefrac}       
\usepackage{microtype}      
\usepackage{graphicx}
\usepackage{doi}
\usepackage{amsmath}
\usepackage{multirow}

\title{Agent-Based modeling in Medical Research - Example in Health Economics}

\date{}

\newcommand{\OpN}[1]{\operatorname{#1}}
\newcommand{\VaN}[1]{\operatorname{\texttt{#1}}}
\newcommand{\VPN}[1]{\operatorname{\texttt{#1}}}
\newcommand{\BDN}[1]{\operatorname{\textsc{#1}}}

\author{
\and
{\large  Philippe Saint-Pierre$^1$},
 \
{\large Romain Demeulemeester$^2$}, \
{\large  Nad\`ege Costa$^3$},\
{\large   Nicolas Savy \thanks{E-mail: \texttt{Nicolas.Savy@math.univ-toulouse.fr}} $^{,1}$}
}

\renewcommand{\headeright}{}
\renewcommand{\undertitle}{}
\renewcommand{\shorttitle}{Agent-Based modeling in Medical Research}

\hypersetup{
pdftitle={Agent-Based modeling in Medical Research},
pdfauthor={Philippe Saint-Pierre},
pdfkeywords={Agent-based model - Medical research - Multivariate distributions - In silico clinical trials - Health economics},
}

\begin{document}
\maketitle

\vspace{-1.0cm}

\begin{center}
$^1$Toulouse Mathematics Institute ; UMR5219 - University of Toulouse ; CNRS  - UPS IMT, Toulouse, France\\[2ex]

$^2$University of Toulouse, Health economic evaluation unit, Medical Information Department, University Hospital of Toulouse, UMR 1295 Centre d'Epid\'emiologie et de Recherche en sant\'e des POPulations, National Institute for Health and Medical Research, Toulouse, France\\[2ex]

$^3$Health economic evaluation unit, Medical Information Department, University Hospital of Toulouse, UMR 1295 Centre d'Epid\'emiologie et de Recherche en sant\'e des POPulations, National Institute for Health and Medical Research, Toulouse, France\\[2ex]
\end{center}

\vspace{1.0cm}

\begin{abstract}
This chapter presents the main lines of agent based modeling in the field of medical research. The general diagram consists of a cohort of patients (virtual or real) whose evolution is observed by means of so-called evolution models. Scenarios can then be explored by varying the parameters of the different models. This chapter presents techniques for virtual patient generation and examples of execution models. The advantages and disadvantages of these models are discussed as well as the pitfalls to be avoided. Finally, an application to the medico-economic study of the impact of the penetration rate of generic versions of treatments on the costs associated with HIV treatment is presented.
\end{abstract}


\section{Introduction} \label{intro}

"Simulation is nowadays consider to be the third pillar of science, a peer alongside theory and experimentation" \cite{S2014}. Indeed, simulation is a relevent way to analyse complex systems and sometimes the only way to analyse such systems. Life science is no exception to this observation even if things are not so simple due to the great variability and the interdependence of the factors involved. A wide range of questions may be investigated by means of simulation: PK-PD issues, translational medicine issues, trial's design optimization issues, sensitivity analysis issues, purely simulation issues (Digital Twin, Simulated Placebo arm)...\\

By simulation approach, the very first idea is to consider a compartmental definition of the virtual patients. By compartmental definition one means top–down model of human from submodels of whole organs to individual molecules. For example the HumMod project  involves more than $1500$ equations and $6500$ variables (body fluids, circulation, electrolytes, hormones, metabolism, and skin temperature,...) (\url{http://hummod.org/}). However, such complex models are very hard to handle and - even with such a large number of variables involved - often too simple essentially. Indeed, the strong dependence structure between variables, pillar of life diversity, is too difficult to take into account.
 
In contrast, agent-based model (ABM), based on numerical simulation may be consider \cite{ABS14,ABS06,ABS13}. There is a wide range of applications of agent-based model \cite{ABM19}. Notice that a standard protocol for describing an ABM has been published in \cite{GRIMM2006115}. The applications to Biology \cite{ABM09} and to Economics \cite{ABM06} are numerous while the applications to medical research are much less numerous. Let us cite the applications to drug development \cite{R13} or to public health \cite{ABMPOL11}. The main lines of the simulation splits into two phases: First, a group of patients (may be real patients  or virtual patients \cite{SSAM}) is considered. Second, the behavior over time of outcomes of each patient is simulated under predefined scenario(s) by means of predictive models usually called execution models \cite{pmid20613720}. The construction of the execution models for outcomes modeling is a pretty technical job involving modeling techniques, databases for calibration of models, expert medical knowledge for variables selection and disease evolution. The agent-based approach yields to results expressed as a distribution of the outcomes from which it is easy to derive prediction intervals at a given risk whereas the population-based approach yields to a punctual prediction of the outcome. Section \ref{S-SCHEMA} is devoted to the description of agent based model in the context of clinical trials.\\

The introduction of simulation techniques in clinical research is probably one of the more relevent way to improve the clinical trial at several steps: design optimization, drug development, clinical operations optimization, monitoring,... The strategy, usually nammed In Silico Clinical Trials (ISCT), aims to identify, by simulation (in silico), design weaknesses, to measure the performance of a trial in a predefined setting while reducing the number of logistical barriers. The purpose being to make the most rational decisions possible regarding clinical development (see \cite{Savy2019} and references). The challenge of In Silico Clinical Trials, an application of agent based model to drug development, is introduced in section \ref{S-ABMISCT}. \\

To construct an agent-based model, two steps have to be investigated: first, to define the cohort of patients to follow up and second to define the execution models involved. The Baseline Data cohort consists in the baseline values of a cohort of patients and concretized as a set of covariates. There are two ways to construct that cohort: by considering an existing cohort or by generating virtual patients. Our attention focuses on the generation of virtual patients. In this context several virtual baseline generators, which consists in Monte-Carlo generations of vectors of covariates, has been developped: discrete method, continuous method (see \cite{pmid17053984} and references therein) and copulas method \cite{bedford2002vines}. These methods are presented and compared in Section \ref{S-VBG}.\\ 

Whatever the objective of the study is, the common line is to the use of exogeneous datasets to learn / calibrate execution models which are used to feed an agent-based model of a (some) pre-specified outcome(s) of interest. An execution model is an Input / Output model which aims to simulate various aspects of the course of the virtual clinical study. There is a wide variety of models available as candidate for execution model: parametric models such as Markov Process, Cox process, regression model, Bayesian network,... and non-parametric models such as machine learning techniques (Random forest, XGBoost, Decision trees, SVM, deep learning,...). Development of execution model is a research question in itself. The question of the prediction error is of paramount importance because the finality of the execution model is to get simulated data and not only to get predicted data. This issue will be discussed in Section \ref{S-ERRPRED}. Examples of execution models in the context of health economics are given in the case study presented in Section \ref{S-AEH}. \\

The paper is organized as follows: Section \ref{S-SCHEMA} is devoted to the main ideas of the construction of agent-based model in the context of clinical research. Section \ref{S-ABMISCT} specifies some properties of ABM for drug development. Section \ref{S-VBG} proposes several strategies for generating cohort of  virtual patients and compare these strategies on a toy example. Section \ref{S-ERRPRED} focuses on the issue of the error of prediction in execution models and investigate the importance of accounting for this error on the toy example. Section \ref{S-AEH} give the main lines of a study \cite{D21} on VIH treatments switching to generics involving an ABM. Finally, Section \ref{S-RECO} states some conclusions and recommendations on the use of ABM in medical research.

\section{Agent-Based Modeling (ABM) in medical research} \label{S-SCHEMA}

\subsection{General schema of an ABM}

In the context of medical research, Agent-Based modeling consists in considering a cohort of patients (agents) and mimicking the behavior of these patients in a virtual medical research according to predefined scenarios. An Agent-Based model involves thus:
\begin{itemize}
\item A cohort of patients values at baseline nammed \textbf{Baseline Cohort}. This cohort may be real data (existing cohort) or data simulated by means of a so-called \textbf{Virtual Baseline Generator} (VBG). This model aims to generate a dataset of virtual patients’ covariates stochastically. The constraints on this model can be summarized in two points: to be consistent with the research protocol we aim to investigate, to be well balanced between complexity of the model and realism of the virtual patient. For virtual patients, the VBG has to generate data in such a way that the marginal distributions are consistent with the ones of the population of interest and the correlation structure between covariates is consistent with the one of the population of interest. Details are given in Section \ref{S-VBV}. 
\item Update or/and enrichment of the cohort by means of \textbf{Execution Models}. The evolution of the cohort in time and the completion of the cohort is driven by execution models. Execution models are input/output models which aim to complete or to modify the virtual clinical dataset. Various execution models can be considered. However, some precautions should be taken, in particular the fact that they are simulation models and not only prediction models. It is very important to take into account the prediction error. This point will be discussed in Section \ref{S-VO}.
\end{itemize}

\begin{figure}[htb]
	\begin{center}
		\includegraphics[scale=0.5]{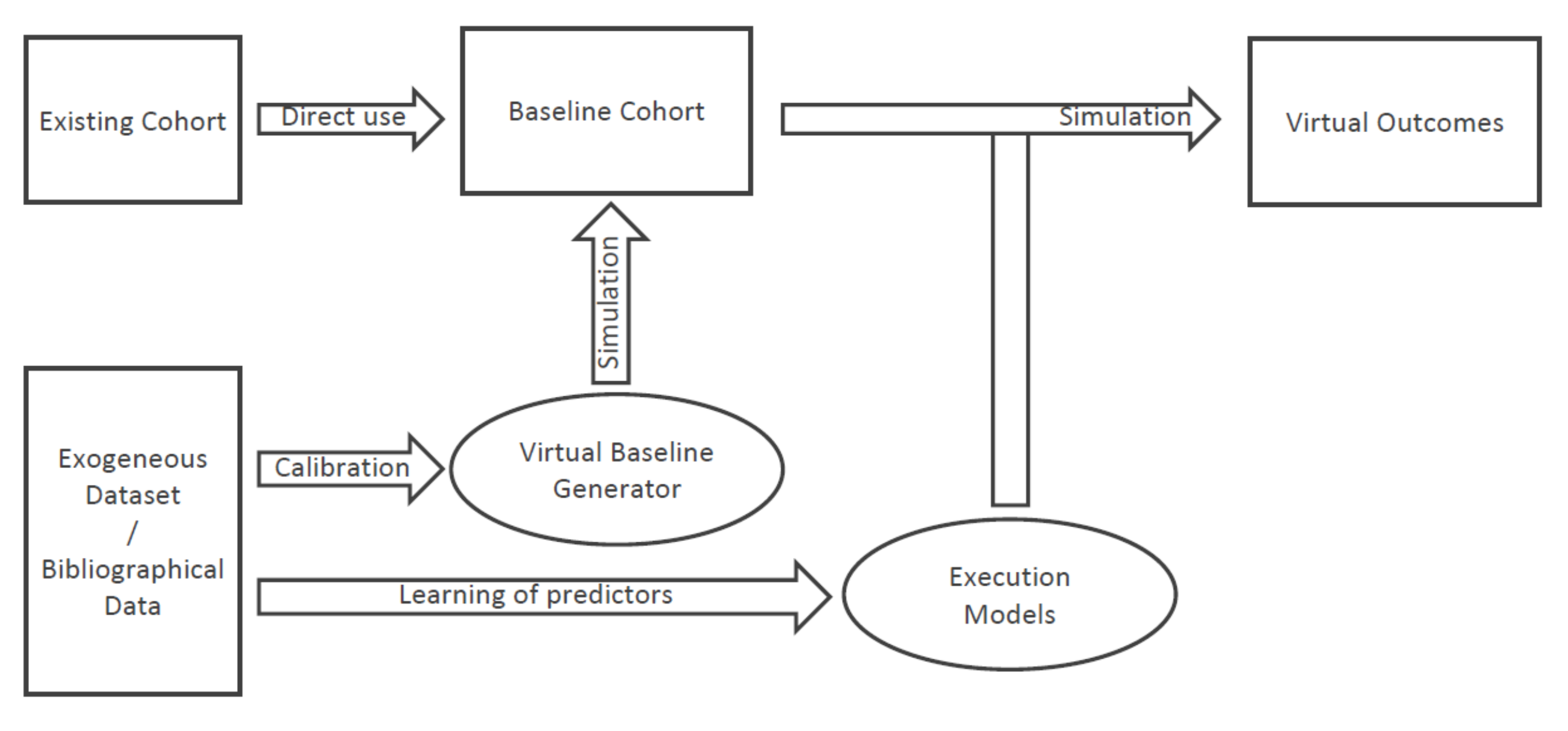} 
		\caption{Example of the simulation design’s schema of an ABM for medical research.} \label{F-GSABM}
	\end{center}
\end{figure}

Agent-Based Models or medical research makes use of exogeneous data to define the VBG and to calibrate / learn the execution models. A generic schema is shown in Figure \ref{F-GSABM}.

\subsection{What an Agent-Based modeling for?}

\subsubsection{Perform sensitivity analyses}

Agent-Based Model allows to perform sensitivity analyses of clinical research endpoints. To do so, it is possible to assess the performance of a study as a function of parameters by varying the feature of the patients, the parameters of the design and parameters of the execution models. Indeed, it is possible to:
\begin{itemize}
\item Modify the feature of the patients by modifying the marginal distributions of the baseline covariates in the virtual baseline generator,
\item To specify experimental design parameters by specifying various scenarios, this means various set of parameters in execution models, and by assessing the related trial performances. 
\item To assess the impact on the trial performances of small changes in the values of the parameters of the execution models.
\item To explore the trial performances for untested values of the parameters of the execution models (for instance what would be the consequences on the performances of a trial in which a patient is followed one year if it is extended to two years).
\end{itemize}

\subsubsection{Perform performances analyses of predefined scenarios}

In order to demonstrate that the difference observed between exposed and unexposed patients is due to the intervention (for example a treatment), an usual way is to assess quantity such as the Average Treatment Effect ($\OpN{ATE}$) defined as:
\begin{equation} \label{ATE}
\OpN{ATE} = \mathbb{E}\left[ Y(1) - Y(0) \right].
\end{equation}
where for patient $i$, $Y_i(1)$ is the outcome for patient $i$ exposed and $Y_i(0)$ is the outcome for patient $i$ unexposed. $Y_i(1)$ and $Y_i(0)$ are potential outcomes \cite{rubin1978} and in practice, both these values cannot be observed simultaneously and $\OpN{ATE}$ cannot be estimated properly. The Average Treatment Effect is usually estimated by
$$
\hat{\OpN{ATE}} = \frac{1}{n_A} \sum_{i=1}^{n_A} Y^A_i -  \frac{1}{n_B} \sum_{i=1}^{n_B} Y^B_i
$$
where $(Y^A_i, i = 1, \dots, n_A)$ (resp. $(Y^B_i, i = 1, \dots, n_B)$) is a sample of patients exposed (resp. unexposed). In the setting of a randomized trial, the quality of this estimation is rather good up to unmeasured confounded factors.

In the context of an Agent-Based Model, virtual patients can explore several arms (or real patients can perform another arm in the context of "digital twin". The performances of this predefined trial can thus be assessed properly since $\OpN{ATE}$ can be estimated directly from \eqref{ATE} by:
$$
\hat{\OpN{ATE}} = \frac{1}{n} \sum_{i=1}^{n} ( Y_i(1) - Y_i(0) ).
$$

\subsection{Baseline Cohort} \label{S-VBV}

The constitution of a cohort of patients who will be involved in the virtual research is the first step of an Agent-Based model. To do so, essentially two strategies may be followed: to make use of an existing cohort of patients or to make use of a Virtual baseline generator to simulate a virtual cohort of patients. Both methods have their pros and their cons discussed above.

\subsubsection{Use of an existing cohort}

The main advantage of working with an existing cohort is that it is realistic, there is no bias inherent in the construction of the cohort. On the other hand, it is not possible to modify the population of interest, in particular to extend the latter or to specified a subgroup without reducing the sample size. Such an approach is perfectly relevant for "digital twins" studies in which the patients of the existing cohort are involved in a virtual research.

\subsubsection{Use of a Virtual Baseline Generator (VBG)}

The advantage of using a virtual cohort are the disavantages of using an existing cohort and the opposite too. On the one hand thanks to the Virtual Baseline Generator, it is possible (and easy) to increase the sample size, to specify a subpopulation or to extend the population. On the other size, the quality and realism of the cohort depends of the performances of the VBG.\\
 
Virtual Baseline Generator involves essentially Monte Carlo generation of a vector of covariates (see, for instance, \cite{RC04,RC10} for details). The issues linked to VBG are the usual ones of Monte Carlo generation for multivariate distributions (see \cite{sJOH87a}) and are linked to the balanced to be found between details of the virtual patients and complexity of the model involved to generate such patients. The problem is magnified by the type of distributions involved which can mix categorical and quantitative variables. The two constraints on the model are: the marginal distributions have to be consistent with the ones of the population of interest and the correlation structure between covariates has to be consistent with the one of the population of interest. Essentially three approaches are mentionned in the litterature : Discrete's method \cite{pmid17053984,Savy2019}, Continuous's method \cite{pmid17053984,Savy2019} and Copula's method \cite{bedford2002vines}.  Description of these methods together with illustration of their use are given in Section \ref{S-VBG}.

\subsection{Virtual Outcomes} \label{S-VO}

\begin{figure}[htb]
	\begin{center}
		\includegraphics[scale=0.5]{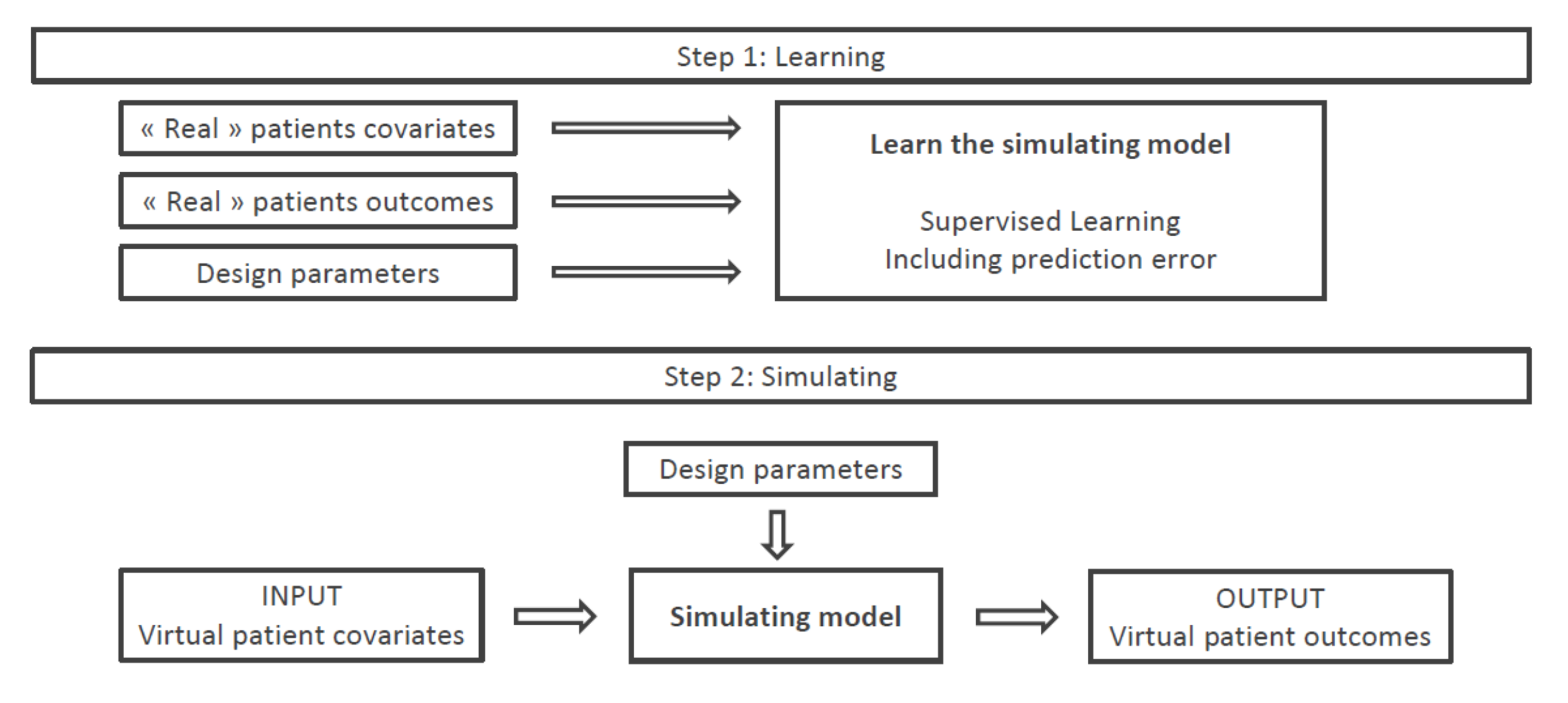} 
		\caption{General steps of an Execution model.} \label{F-EXECMOD}
	\end{center}
\end{figure}

As discussed above, an execution model is an Input / Output model which aims to simulate the course of the virtual clinical trial. The building splits in two step. First, a learning step involving a set of (real) patients covariates together with the outcome of interest (learning database). A predictive model of the outcome of interest is learnt (calibrated) from this learning database. The second step consists in the simulation of the outcome from the (virtual or real) patients of interest thanks to the model learnt at step 1. Figure \ref{F-EXECMOD} schematizes a general execution model. \\

Many execution models can be considered involving many learning models from linear regression to machine learning through Markov processes. To insure the versatility of the construction of an ABM, each model may be improved separately. This property is known as the modularity and is a property of major importance for ABM building and improvement. It is important to note that those models depend on parameters that can be split in three categories: parameters linked to the patients, parameters linked to the model (tuning parameters) and parameters linked to the design. Those parameters can be considered as punctual (deterministic) values or as distributions (random) in a Bayesian's paradigm. Parameters can be fixed by the user or estimated from databases. Parameters fixed by the user, by means of a Human Machine Interface, state a scenario which define the conditions under which the trial is followed. Parameters estimated from databases are calibration parameters which are fixed during the simulation.

\section{ABM and drug development: In Silico Clinical Trials (ISCT)' challenge} \label{S-ABMISCT}

For about thirty years, the use of simulation techniques have been introduced in drug development. As early as 2009, the use of simulation techniques proved their benefits. In \cite{pmid19327954} Brindley and Dunn shown that simulations studies increase the probability of achieving objectives of the study, increase patient safety, reduce the duration of the study and the risk of protocol deviations and avoid inconclusive situations. The introduction of simulation in clinical research seems natural but the literature does not confirm this idea. Indeed, the state of the art \cite{pmid10836134} relating to the period prior to 2000 and confirmed by the reviews \cite{pmid20613720} over the period 2000-2010 and \cite{SSAM} over the period 2010-2015 show little impact and use. Explanations of this paradox may be found in reporting bias, as such investigation may be conducted by pharmaceutical companies and not necessarily published for confidentiality reasons.\\

It is important to emphasize that the regulation agencies are boosting the use of simulation in drug development \cite{BL4}. As an example, in 2011, the FDA released its strategic plan on Advancing Regulatory Science (\url{https://www.fda.gov/media/81109/download}). Four of the eight science priority areas evoked specifically call out modeling and simulation as important aspects of FDA’s strategy. In the FDA Grand Rounds presentation of Dr. Tina Morrison (Chair of FDA’s Agency-wide Modeling and Simulation Working Group and Regulatory Advisor of Computational Modeling for FDA’s Office of Device Evaluation) given on August 9, 2018 an overview of some current modeling and simulation methodologies were provided and the potential of in silico clinical trials were discussed (\url{https://collaboration.fda.gov/p4r7q3qweuv/}). Another example, during the last 10 years the European Medicines Agency (EMA) organized a number of workshops on modeling and simulation, working towards greater integration of modeling and simulation (M\&S) in the development and regulatory assessment of medicines. In the 2011 EMA - EFPIA (European Federation of Pharmaceutical Industries and Associations) Workshop on Modelling and Simulation, European regulators agreed to harmonize on good M\&S practices and for continuing dialog across all parties. To do so, the EMA Modelling and Simulation Working Group (MSWG) has been established and the MID3 (Model-Informed Drug Discovery and Development) good practices paper publisehd in 2016 \cite{pmid27069774,pmid28653481}.\\

\begin{figure}[h!]
	\begin{center}
		\includegraphics[scale=0.6]{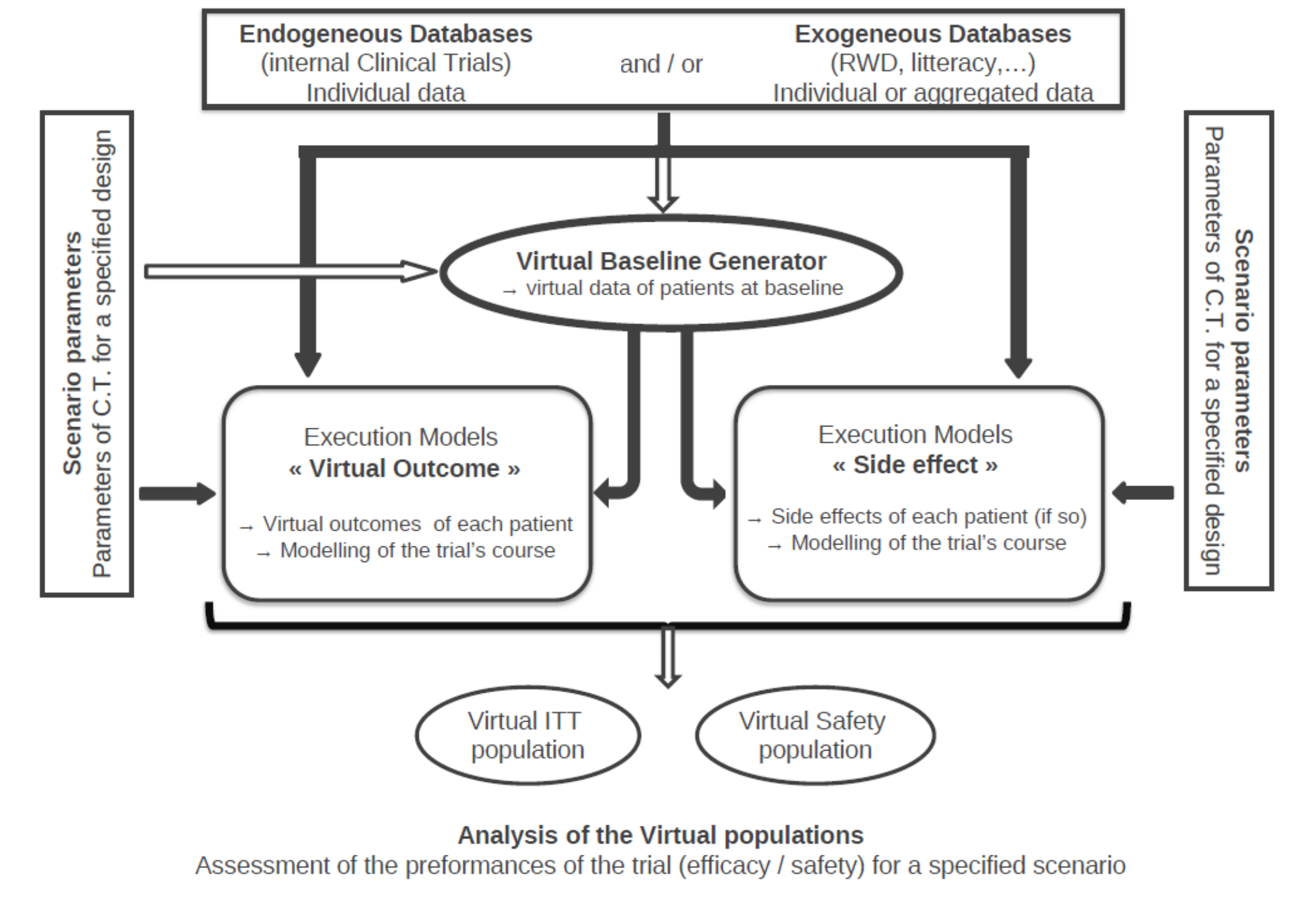} 
		\caption{Example of the simulation design’s schema of an ISCT.} \label{F-SCHEMA}
	\end{center}
\end{figure}

Agent-Based modeling in clinical research concretizes by "In Silico Clinical Trials", ISCT for short, which consists in the use of patients (virtual or real) to mimic their behavior in a virtual clinical trial in order to challenge trial’s design in terms of feasibility and probability of success of the trial. The main idea is to use the huge amount of information available on the patients, on the drug of interest and on the design of the trial in order to build a stochastic model mimicking the course of a clinical trial. In the context of ISCT, (virtual) patients is defined as a set of covariates essentially the ones involve in the clinical design (inclusion / exclusion criteria) and the ones known to be linked with the clinical outcomes of interest. Whatever the objective of the study is, the common line is to the use of exogeneous datasets to learn / calibrate a predictive model which is used to feed an agent-based model of a (some) pre-specified outcome(s) of interest. As examples of execution models one finds Baseline's parameters evolution in time (an example is given in \cite{D21} and in Section \ref{S-AEH}), Virtual outcome generator, Disease progression model \cite{pmid28936389}, Side effect model, drop-out model, Patients Recruitment model \cite{MSS12,pmid29740630}. There is a wide variety of models available as candidate for execution model: parametric models such as Markov Process, Cox process, regression model, Bayesian network,... and non-parametric models such as machine learning techniques (Random forest, XGBoost, Decision trees, SVM, deep learning,...). Development of execution model is a research question in itself. Figure \ref{F-SCHEMA} is an example of what may be the simulation's schema of an ISCT involving an execution model which complete the dataset with virtual outcomes at different times and an example of execution model which modifies the dataset introducing adverse events.\\ 

ISCT allows to perform sensitivity analyses of clinical trial endpoints by assessing the performance of the trials as a function of parameters by varying the feature of the patients, the parameters of the design and parameters of the execution models.
\begin{itemize}
\item To modify the feature of the patients relies to challenge the inclusion / exclusion criteria of a trial.
\item To specify design parameters like duration of patients' follow-up, number of centers involved... 
\item To assess the impact on the trial performances of small changes in the values of the parameters of the execution models for instance to quantify the impact of a given variation of patients' recruitment rate on the trial duration.
\item To explore the trial performances for untested values of the parameters of the execution models.
\end{itemize}
It is important to point that these two last strategies are much more easy to investigate with parametric execution models than non-parametric ones.

\section{Virtual Baseline' generator} \label{S-VBG}

If the joint distribution of the covariates, denoted $f_{(C^1, \dots, C^K)}$ for simplicity, is known, the strategy is nothing but a Monte Carlo generation of data. To be meaningful, the data generation has to rely with available informations on the covariates. That information may comes from the literature (bibliographical data) or from an existing database (historical database). Dealing with bibliographical data, the covariates distributions parameters are parameters of the population while dealing with historical data, these parameters are estimations of population parameters.  Bibliographical data are more relevant but, most of the time, only marginal distributions or resume of marginal distributions can be obtained. The better we can do is often to assume covariates as independent, that is not satisfactory. From historical database, whole the correlation structure may be estimated but the question of how to estimate  parameters with enough precision raises. In what follows, the question of reconstruction of a database  is to be understood in the sense on how to generate a database with marginal distributions and the correlation structure close to the original ones.\\

In what follows, $C^k$ denotes the $k$-th covariate ($k$ varies from 1 to $K$) and $c^k_i$  denotes the values of that covariate of the $i$-th patient ($i$ varies from 1 to $n$). For a sake of notational simplicity $\vec{c}_i$ denotes  the vector of patient $i$ covariates' values. Covariate $C$ may  be continuous (denoted $^cC$) or categorical (denoted $^dC$). Throughout that section, continuous covariates are assumed to be normally distributed but it can be any other distribution up to a change of variable.

\subsection{Discrete method}

Discrete Method consists in spliting continuous variables and discrete variables by conditioning, writing, with notational abuse:
$f_{(C^1, \dots, C^K)} = f_{((^cC^1, \dots, ^cC^L)| (^dC^{L+1}, \dots, ^dC^K))} \times f_{(^dC^{L+1}, \dots, ^dC^K)} $. 

\subsubsection{Calibrating method}

The calibration of the model splits in two steps:
\begin{itemize}
\item First, fit the distribution of  $(^dC^{L+1}, \dots, ^dC^K)$ by estimating the proportion of each modality,
\item Second, fit the distribution of $f_{((^cC^1, \dots, ^cC^L)| (^dC^{L+1}, \dots, ^dC^K))},$ estimating the mean vector and variance-covariance matrix for each configuration $(^dC^{L+1}, \dots, ^dC^K)$ .
\end{itemize}

\subsubsection{Generating dataset}

The simulation of a database of size $n$ consists in performing $n$ times the algorithm
\begin{itemize}
\item Draw a configuration with the probabilities estimated at step 1,
\item Given this configuration, draw the remaining values from the  multinormal distribution estimated at step 2.
\end{itemize}
Notice that the problem can be split in groups of independent covariates. In situation where the parameters of the distributions are known, the Discrete method is exact. But most of the time these parameters are estimated from data and Discrete method is less efficient especially when there is a large number of covariates mixing continuous and categorical ones. Indeed this technique needs a lot of estimations and, for situations where there are a lot of categorical covariates, the estimation of the parameters of the continuous covariates is poor because of a small number of data for several modalities. 

\subsection{Continuous method}

The Continuous method is introduced in \cite{pmid17053984} to generate database directly from the population parameters. This approach consists in considering all the covariates as normally distributed.

\subsubsection{Learning / Calibrating method}

Calibration step consists in fitting a multinormal distribution for whole the covariates whatever the type of covariates involved. For categorical variable, a recoding of variable may be used in order to consider only numerical data.

\subsubsection{Generating dataset}

The simulation of a database of size $n$ consists in drawing $n$ values $(u_i^1, \dots, u_i^K)_{i=1, \dots,n}$ from the multinormal distribution $N(\vec{\mu},\Sigma)$ estimated in the learning step.
\begin{itemize}
\item For a continuous covariate $^c c^k = u^k$ for $k=1,\dots,L$
\item For a categorical covariate  with $M$ modalities $^d C^k$, one makes use of the critical values defined by
$$
\begin{cases}
CrV^k_m = \mu_k + \Sigma_{k,k} \phi^{-1}( \sum_{i=1}^m p_i), &\qquad 1 \leq m \leq M-1 \\
CrV^k_M = + \infty \\
CrV^k_0 = - \infty 
\end{cases}
$$
where $(p_i;~1 \leq i \leq M)$ are the proportions of each modality of the categorical covariate, $\mu_k$ and $\Sigma_{k,k}$ are the parameters of the  normal distribution, $\phi$ denotes the cumulative function of standard normal distribution.\\
Finally, $^d c^k = m$ if and only if $CrV_{m-1}^k < u_i^k \leq CrV_m^k$.
\end{itemize}

The Continuous method is an appealing alternative even if the multivariate Gaussian assumption may be to restrictive especially to catch multi-modal distributions. Comparison of Discrete and Continuous methods given bibliographical data has been investigated in \cite{pmid17053984} assuming continuous covariates normally distributed. In this context the calibration step is out of purpose. In \cite{SSAM,Savy2019} attention is paid on the performances of these techniques on a closer to practice setting,  when a historical database is given. The aim of the machinery  is to generate a realistic copy of this historical database (the marginal distributions coincides and the correlation structure between covariates is preserved). 


\subsection{Copula method}

\subsubsection{Copula}

Copula approach is a practical and powerful tool to construct multivariate distributions. A copula aim to describe the dependence structure between a group of random variables and the specification of copulas can be done independently from the marginal distributions. Formally, a copula is a multidimensional cumulative distribution function (CDF) linking the margins of the vector of random variables $\mathbf{X} = (X_1, \dots, X_d) \in \mathbb{R}^d$ to its joint distribution. Let $F_i(x_i) = \mathbb{P} [X_i \leq x_i]$ the continuous CDF of $X_i$ for $i=1,...,d$, the Sklar's Theorem \cite{Sklar96} state that every multivariate distribution $F_{\theta}$ can be written as

\begin{equation}
F_{\theta} (\mathbf{x}) = \mathcal{C}_{\theta} \left(F_1 (x_1), \dots, F_d (x_d) \right),
	\label{eq:copula_function}
\end{equation}

for some appropriate $d$-dimensional copula $\mathcal{C}_{\theta}$ with parameter $\theta \in \Theta$. If all marginal distributions are continuous functions, then there exists a unique copula satisfying

\begin{equation*}
\mathcal{C}_{\theta} \left(u_1, \dots, u_d \right) = F_{\theta} \left( F^{-1}_1 (u_1), \dots,F^{-1}_d (u_d) \right)
\end{equation*}

where $u_j = F_j(x_j)$. For $F_{\theta}$ absolutely continuous with strictly increasing marginal distributions, one can derive the joint density of $\mathbf X$ from \eqref{eq:copula_function}:

	\begin{equation*}
		f_\theta (\mathbf{x}) = \left[\prod_{i=1}^d f_k(x_k) \right]   c_\theta \left(F_1 (x_1), \dots, F_d (x_d) \right),
		\label{eq:copula_density}
	\end{equation*}
where $c_\theta$ denotes the copula density and $f_i(x_i)$, $i=1,...,d$, correspond to the marginal densities.

Many parametric copula families are available and are based on different dependence structures \cite{Nelsen07}. Most of these families have bi-dimensional dependencies even if some can be extended to higher dimensions. A first and naive approach for modeling the dependence structure is assuming a Gaussian copula for this last. In this case, the problem is reduced to determining the input correlation matrix. Such multivariate Gaussian assumption is however very restrictive. Multivariate Archimedian copulas can also be used to describe asymmetric tail dependencies.  However, the same copula family describes the dependence among all the pairs, which is not flexible enough. Since the choice of copula has a strong impact on the output distribution, we will consider, in the following, a flexible approach setting by modeling the input distribution using regular vine copulas (R-vines). R-vine copulas \cite{Joe94} can overcome the constraint of bi-dimensionality and the lake of flexibility of multidimensional copulas by describing multidimensional dependencies by combining multiple pair-copulas.

\subsubsection{R-vine Copula}

Representing multi-dimensional dependence structures in high dimensional settings is a challenging problem. A probabilistic construction of a multivariate distribution function based on pair-copulas has been developed by \cite{joe1996families}. It offers a very flexible way to construct high-dimensional copulas, but is not unique. For describing  all such possible constructions in an efficient way, \cite{bedford2001probability} have introduced the regular vines (R-vines). This graphical tool, based on a sequence of trees, gives a specific way to decompose the multivariate probability distribution. 

For the following definition, we omit the use of $\theta$ in $f_{\theta}$, $F_{\theta}$ and $c_{\theta}$ for a sake of simplicity. The Sklar's theorem states that a copula $C$ is unique and admits a density $c$ if the margins $F_1, \dots, F_d$ are continuous. Using the chain rule, the joint density $f$ associated to the random vector $\mathbf{X}=(X_1,\dots,X_d)$ can be written as 
	\begin{equation}
		f(x_1, \dots, x_d) = c(F_1(x_1), \dots, F_d(x_d)) \times \prod_{i=1}^d f_i(x_i),
				\label{eq:CopDensity}
	\end{equation}
	where $f_i(x_i)$, $i=1,...,d$, correspond to the marginal densities. By recursive conditioning, the joint density can also be written as

 	\begin{equation}
		\begin{split}
		f(x_1, \dots, x_d) &= f_d(x_d) \times f_{d-1/d}(x_{d-1} | x_d) \times f_{d-2/d-1,d}(x_{d-2} | x_{d-1}, x_d) \\ 	 
		&\times \cdots \times f_{1/2,\dots,d}(x_1 | x_2, \dots, x_d),		
		\end{split}
		\label{eq:PCC_eq1}
	\end{equation}
which is unique, up to a re-labelling of the variables.

For example, in a case of three variables, one possible decomposition of the joint density
can be written as	
	\begin{equation}	
	f(x_1,x_2, x_3) = f_1(x_1) \times f_{2/1}(x_{2} | x_1) \times f_{3/2,1}(x_{3} | x_{2}, x_1)
		\label{eq:ExDim3}
	\end{equation}	

The conditional densities can be rewritten using Bayes formula 

    \begin{equation}
    		\begin{split}
		f_{3/1,2}(x_3|x_1, x_2) &= \frac{f_{3,1/2}(x_3,x_1|x_2)}{f_{1/2}(x_1|x_2)} \\
		&=\frac{ c_{1,3|2}(F_{1/2}(x_1|x_2), F_{3/2}(x_3|x_2)) \times f_{1/2}(x_1|x_2) \times f_{3/2}(x_3|x_2)  }  { f_{1/2}(x_1|x_2) }  \\
		&= c_{1,3|2}(F_{1/2}(x_1|x_2), F_{3/2}(x_3|x_2)) \times f_{3/2}(x_3|x_2)
		    \end{split}		
		\label{eq:PCC_3d}
	\end{equation}

	where the second equality comes from Sklar's theorem. By developing $f(x_3|x_2)$ in \eqref{eq:PCC_3d} in the same way we find that
	\begin{equation}
		f_{3/1,2}(x_3|x_1, x_2) = c_{1,3|2}(F_{1/2}(x_1|x_2), F_{3/2}(x_3|x_2)) \times c_{2,3}(F_{2}(x_2), F_{3}(x_3)) \times f_3(x_3)
		\label{eq:ExDim3_1}		
	\end{equation}
By doing the same procedure for $f(x_2|x_1)$ in \eqref{eq:ExDim3}, one can derive a pair-copula decomposition for the joint density
	
	\begin{equation}	
	    		\begin{split}
	f(x_1,x_2, x_3) &= f_1(x_1) \times f_2(x_2) \times f_3(x_3)  \\    
	                            &\times c_{1,2}(F_{1}(x_1), F_{2}(x_2))
	                             \times c_{2,3}(F_{2}(x_2), F_{3}(x_3)) \\
	                              &\times c_{1,3/2}(F_{1/2}(x_1/x_2), F_{3/2}(x_3/x_2))	
	           \end{split}		                      	
		\label{eq:ExDim3_2}
	\end{equation}
	
Such representation of the joint density in terms of pair-copulas and marginal density is called the pair-copula construction (PCC). The resulting decomposition offers a very flexible way to construct high-dimensional copulas and to model flexible dependence structures using only bivariate copulas. However, the decomposition is not unique. Indeed, \eqref{eq:CopDensity} has numerous decomposition forms and it increases significantly with  $d$. 

\begin{figure}
		\centering
		\includegraphics[width=0.5\textwidth]{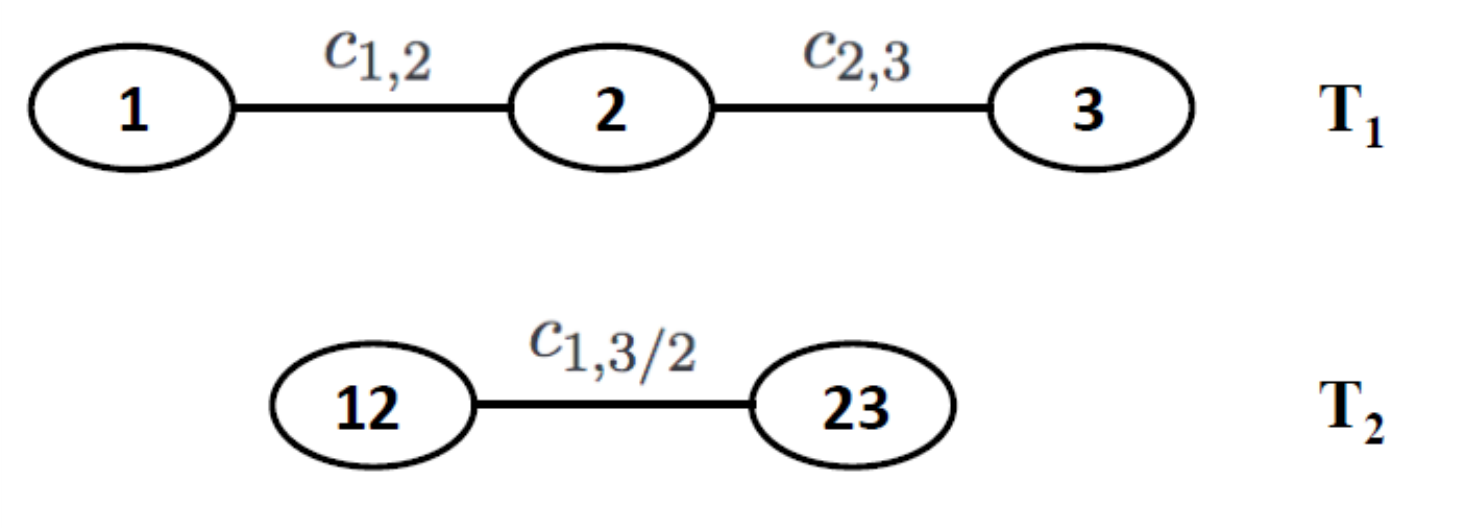}
		\caption{R-vine structure of a dimension $d=5$ problem.}
		\label{fig:R_vine_example_5d}
\end{figure}

To help organising them, a graphical model called regular vine (R-vine) was introduced in \cite{bedford2001probability, bedford2002vines} and detailed in \cite{kurowicka2006uncertainty}. Each of these sequences of tree gives a specific way of decomposing the density. A R-vine describe a $d$-dimensional PCC and is a sequence of linked trees where the nodes and edges correspond to the $d(d-1)/2$ pair-copulas. According to Definition 8 in \cite{bedford2001probability}, a R-vine consists of $d-1$ trees $T_1, \dots, T_{d-1}$ with several constraints. Each tree $T_i$ is composed of $d-i+1$ nodes which are linked by $d-i$ edges for $i = 1, \dots, d-1$. A node in a tree $T_i$ must be an edge in the tree $T_{i-1}$, for $i=2, \dots, d-1$. Two nodes in a tree $T_i$ can be joined if their respective edges in tree $T_{i-1}$ share a common node, for $i=2, \dots, d-1$. The pair copula in the first tree characterize pairwise unconditional dependencies, while the pair copula in higher order trees model the conditional dependency between two variables given a set of variables. The number of conditioning variables grows with the tree order.  Figure \ref{fig:R_vine_example_5d} illustrated a specific R-vine associated to a density decomposition in dimension 5. Note that a PCC where all trees have a path like structure define the D-vine subclass while the star like structures correspond to C-vine subclass \cite{bedford2001probability}. 

	
\subsubsection{Discrete variables and mixed case}

In practice, the covariates are of mixed nature with discrete and continuous variables. The extension of the PCC to the discrete case has been studied by  \cite{Panagiotelis2012}. Finally, in the discrete case, the calculations are the same as in the continous case using probabilities instead of densities. In this case, techniques based on discrete differentiation should be used instead of derivatives. The analogue of conditional densities are pointwise conditional probabilities. The amplitude of the jumps must be evaluate at all considered values. The mixed case with discrete and continuous variables has been studied in \cite{Stober2015}. The results in the continous and in the discrete case can be combine to derive the pairwise decomposition in the mixed case. 

\subsubsection{Learning / Calibrating method}

The calibration of an R-vine to a multivariate distribution finally consists in 
\begin{itemize}
	\item selecting the tree structure,
	\item selecting a copula family for each pair copula in the decomposition,
	\item estimating the parameter for each pair copula.
\end{itemize}	
In \cite{Aas2009mle} authors have proposed an approach for selection of R-vine model based on maximum likelihood estimation. Sequential estimation \cite{kurowicka2011estimation, dissmann2013selecting}, truncation \cite{aas2012truncated} or bayesian estimation \cite{Gruber2015bayesian} have also been proposed to select the R-vines. Most of the recent developments, in particlular the extension to the mixed case (discrete and continuous variables), are implemented in the library \textit{vinecopulib} available with R and Python. The R package \textit{rvinecopulib} \cite{rvinecopulib} has been used to calibrate the R-vine copula, maximum likelihood estimation with AIC selection criterion has been considered.
	
\subsubsection{Generating dataset}

The simulation of a database of $K$ covariates of size $n$ consists in drawing $n$ values $(u_i^1, . . . , u_i^K)_{i=1,...,n}$ from independent uniform distribution. These uniform marginals are then given to the R-vine model to be transformed to the copula sample.  The $n \times K$ matrix of data simulated from the given R-vine copula model are uniform variables which must be transformed to retrieve their original scales. The empirical distributions of the original data and their pseudo-inverse were used to transform the uniform to their original scales.

\subsection{A toy example}

\subsubsection{Pima Indians Diabetes Dataset}

As an illustrative example, consider an extract of the "Pima Indians Diabetes Dataset" which involves predicting the onset of diabetes within 5 years in Pima Indians given medical details. It is a binary (2-class) classification problem where the outcome of interest is the variable "Diabete" Yes or No. A random subset of 392 patients of "Pima Indians Diabetes Dataset" is considered. The covariates involved are qualitatives and quantitatives:
\begin{itemize}
\item Have you ever been pregnant? (Yes - No)
\item Body mass index ($\leq 25$, $]25;35]$, $>35$)
\item Plasma glucose concentration a 2 hours in an oral glucose tolerance test
\item Diastolic blood pressure (mm Hg)
\item Triceps skinfold thickness (mm)
\item 2-Hour serum insulin (mu U/ml)
\item Diabetes pedigree function
\item Age (year)
\end{itemize}

\subsubsection{Virtual patients simulations}

\begin{figure}
		\centering
		\includegraphics[width=1\textwidth]{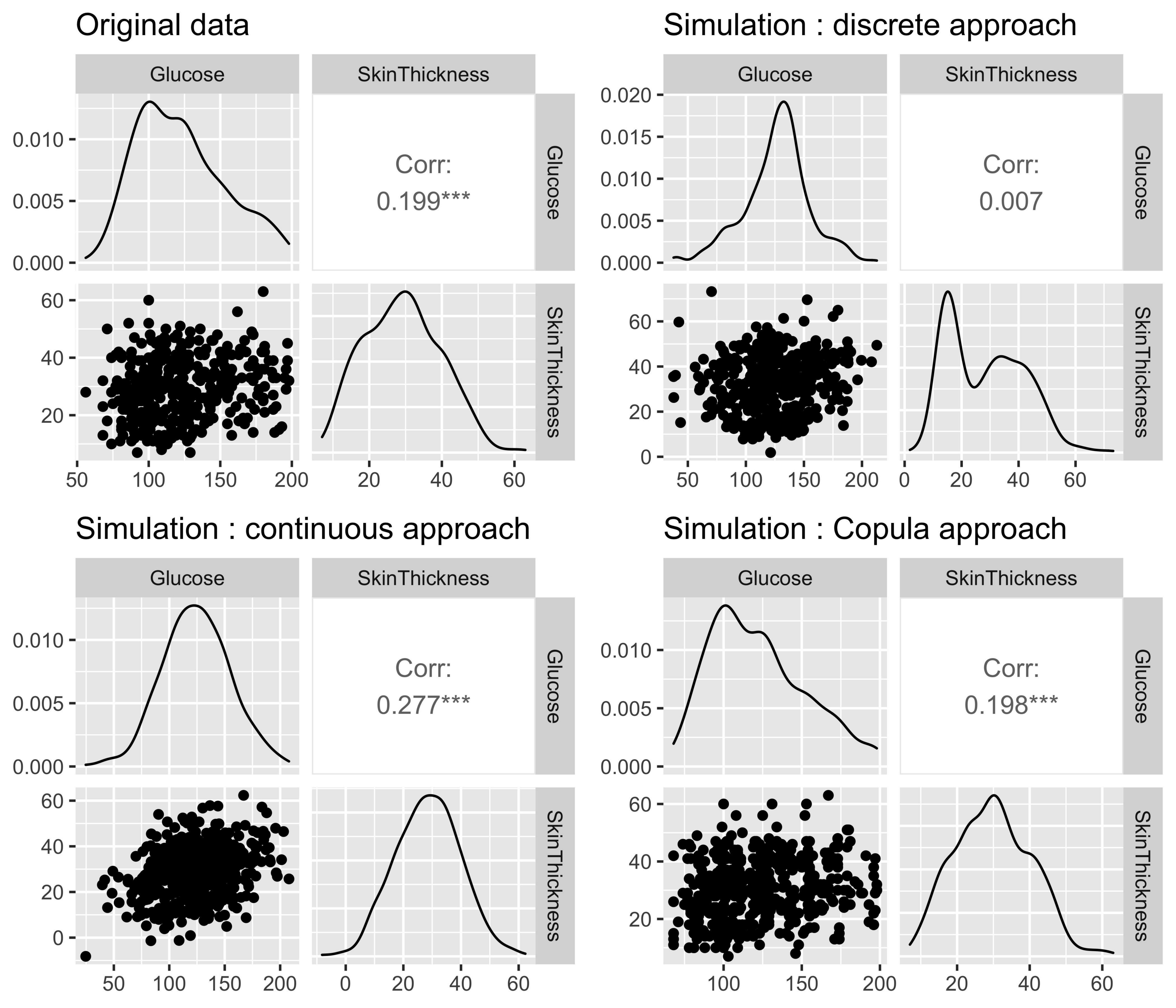}
		\caption{Marginal distribution, correlation and scatter plot of variables \textit{Glucose} and \textit{Skin thickness} for the original data and data simulate using the discrete, the continuous and the Copula approaches.}
		\label{fig:correlation_3vs5}
\end{figure}

\begin{figure}
		\centering
		\includegraphics[width=1\textwidth]{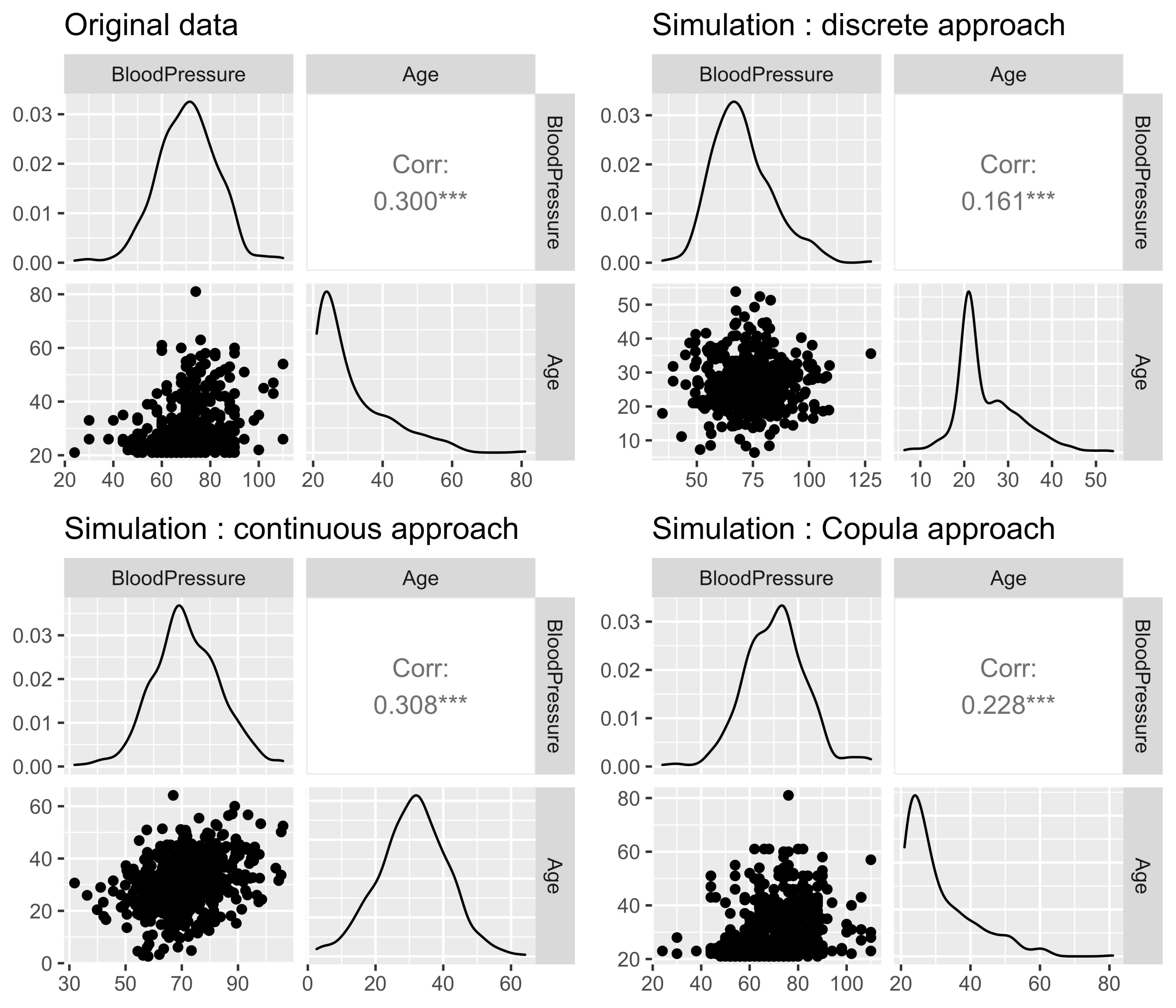}
		\caption{Marginal distribution, correlation and scatter plot of variables \textit{Blood pressure} and \textit{Age} for the original data and data simulate using the discrete, the continuous and the Copula approaches. Simulations are performed by adding or not an error in the prediction of the \textit{Prognostic} variable.}		\label{fig:correlation_4vs8}
\end{figure}

The three simulation approaches can be quickly evaluated graphically by comparing marginal distributions of simulated variables with the marginal distributions of original data. The pairwise correlation between variables and the correlation between variables and prognosis can also be compared to those of the original data. Figures \ref{fig:correlation_3vs5} and \ref{fig:correlation_4vs8} display marginal distributions and correlation for variables \textit{Glucose} and \textit{Skin thickness} and for variables \textit{Blood pressure} and \textit{Age} respectively.  

Figure \ref{fig:correlation_3vs5} seems to indicate that the discrete approach fail to recover the true marginal distribution and the significant correlation between \textit{Glucose} and \textit{Skin thickness}. The continuous approach provides coherent result since the Gaussian assumption seems not to stringent in this application. The Copula approach seems to be the more efficient according to marginal distribution and correlation.

Figure \ref{fig:correlation_4vs8} confirms the conclusion derived from Figure \ref{fig:correlation_3vs5}. Even if the correlation is better recovered with the continuous method, the marginal distribution and the scatter plot are more relevant with the original data. Surprisingly the discrete approach do not performed very well. Indeed, if a covariate is not well balanced, Gaussian parameter estimation may be poor.

\section{Virtual Outcomes} \label{S-ERRPRED}

\subsection{Discussion on the data}

A huge diversity of models may be considered for the execution models: parametric models (Markov, Cox, linear, logistics,...) and non-parametric models (Machine learning). The main difference between those two approaches is parametric models calibrate by means of data and assumptions while non-parametric models are completely data-driven. Many sources of data may be used: completed clinical trials, on-going clinical trials, real-word database with really different levels of quality and levels of accuracy. The quality of the execution model may be very different according to the database used for calibration.

For parametric models, the main issue comes from the data used for parameters estimation. The results are better with large databases since inference is better but if there is no (or not enough) data available it is still possible to make stronger assumptions on the model to simplify it. The other advantage of parametric model is the possibility to use data from literacy, from expert knowledge or fixing a value and perform sensitivity analysis.

For non-parametric models, the main issue comes from the fact that there is no alternative to the data-driven approach. Without data it is not possible to consider a model and with data, the model is much more sensitive to it quantity and to the structure of the database. Indeed, the quality of the prediction is linked with the structure of the database, the model learn from the data explored, if the database do not explore certain values, the prediction associated with such value will be poor or not available. For instance if there is no information in the learning database on young people ($< 30$ years old), the predictor will not predict anything for young people.

Whatever the nature of the model a recurring question is the portability of the data. Indeed, is it realistic to learn a model (parametric or not) from a dataset involving patients completely different of the patients to include in the ABM (for instance learn from American people for an European study). To overpass this problem, various algorithm such as OT-algorithm \cite{gares} may be of interest.

\subsection{Discussion on the generators} 

The output of an execution model comes from a Monte Carlo simulation accounting for the model chosen and for the values of whole the parameters involved. It is important to keep in mind that the aim of an execution model is to simulate an outcome and not only to predict an outcome. That is an important point because model assumptions for simulating are stronger than for predicting. It necessitates a model not only with good predictive performances but also a modelling of the error of prediction. Notice that the confusion between prediction and simulation yields to an under estimation of the error and thus it is easier for a factor effect to be significant.

\subsection{Back to the toy example}

\subsubsection{Virtual outcome simulations}

The outcome of interest in the original data is the binary variable "Diabete" Yes or No. Consider the subgroup of participants satisfying "Diabete = Yes" and the subgroup of participants satisfying "Diabete = No". A model to explain the outcome can be estimated from the original data. This model can then be used to simulate the outcome of interest for the virtual patients. In our case a Random Forest algorithm is learned from the original data. 

A set of $500$ virtual patients are randomly generated by means of the three simulations methods and the corresponding outcome is predicted by means of the random forest algorithm with and without wondering of the error of prediction. In order to take into account the prediction error, the predicted value is generated by means of the confusion matrix of the prediction of the random forest on the original data. For each simulation approaches, $100$ datasets were simulated together with their \textit{Prognostic} variable (the outcome). 

\subsubsection{Comparison of simulation approaches}

\begin{figure}
		\centering
		\includegraphics[width=1\textwidth]{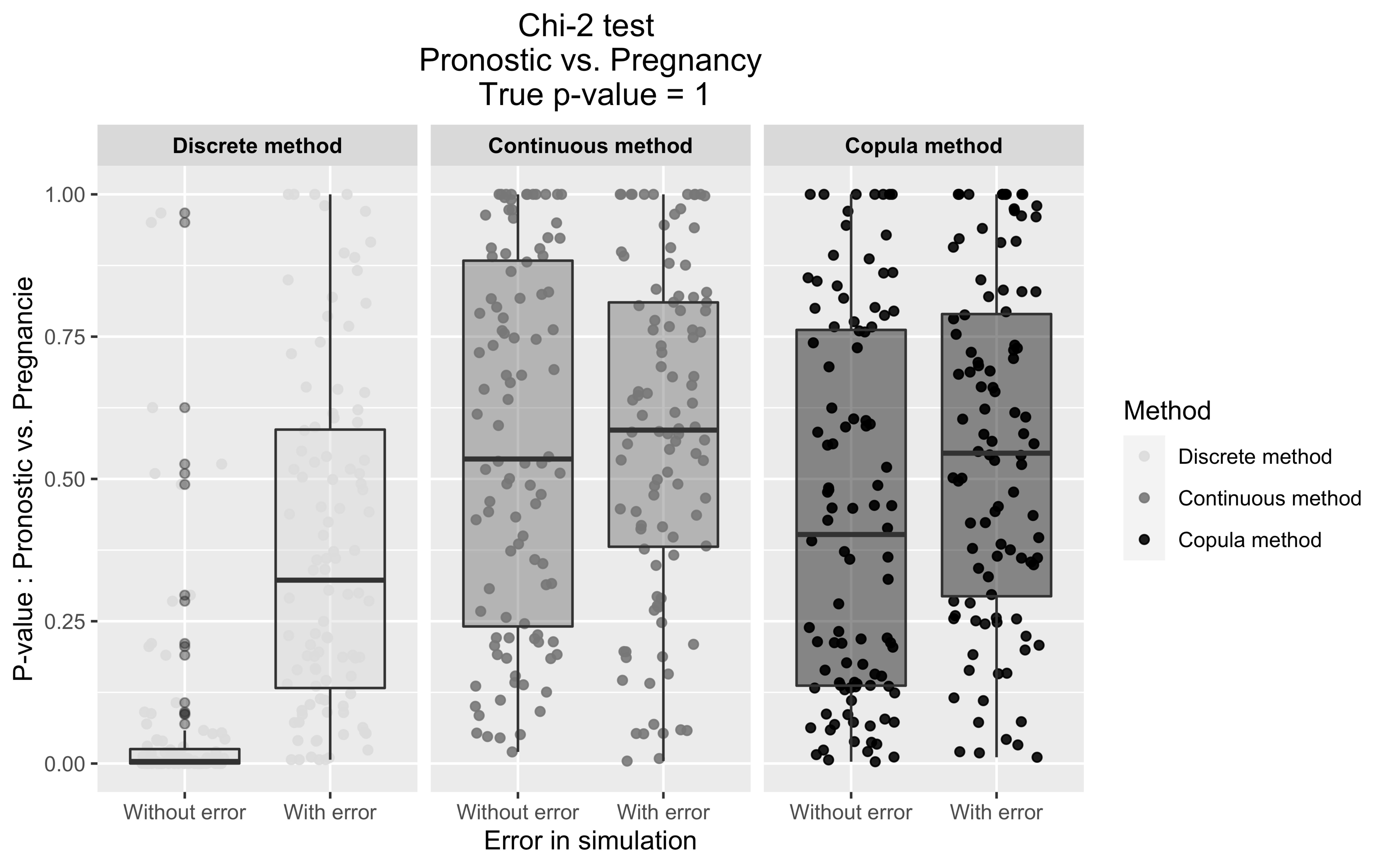}
		\caption{Simulation of $100$ datasets of $500$ observations each were performed using the discrete, the continuous and the Copula approaches. For each dataset, the p-value using Chi-2 test between \textit{Prognosis} variable and \textit{Pregnancy} were computed and reported using boxplot and jitter plot. The p-value observed in the original data is given in the title. Simulations are performed by adding or not an error in the prediction of the \textit{Prognostic} variable.}
		\label{fig:Pvalue_Var1}
\end{figure}

\begin{figure}
		\centering
		\includegraphics[width=1\textwidth]{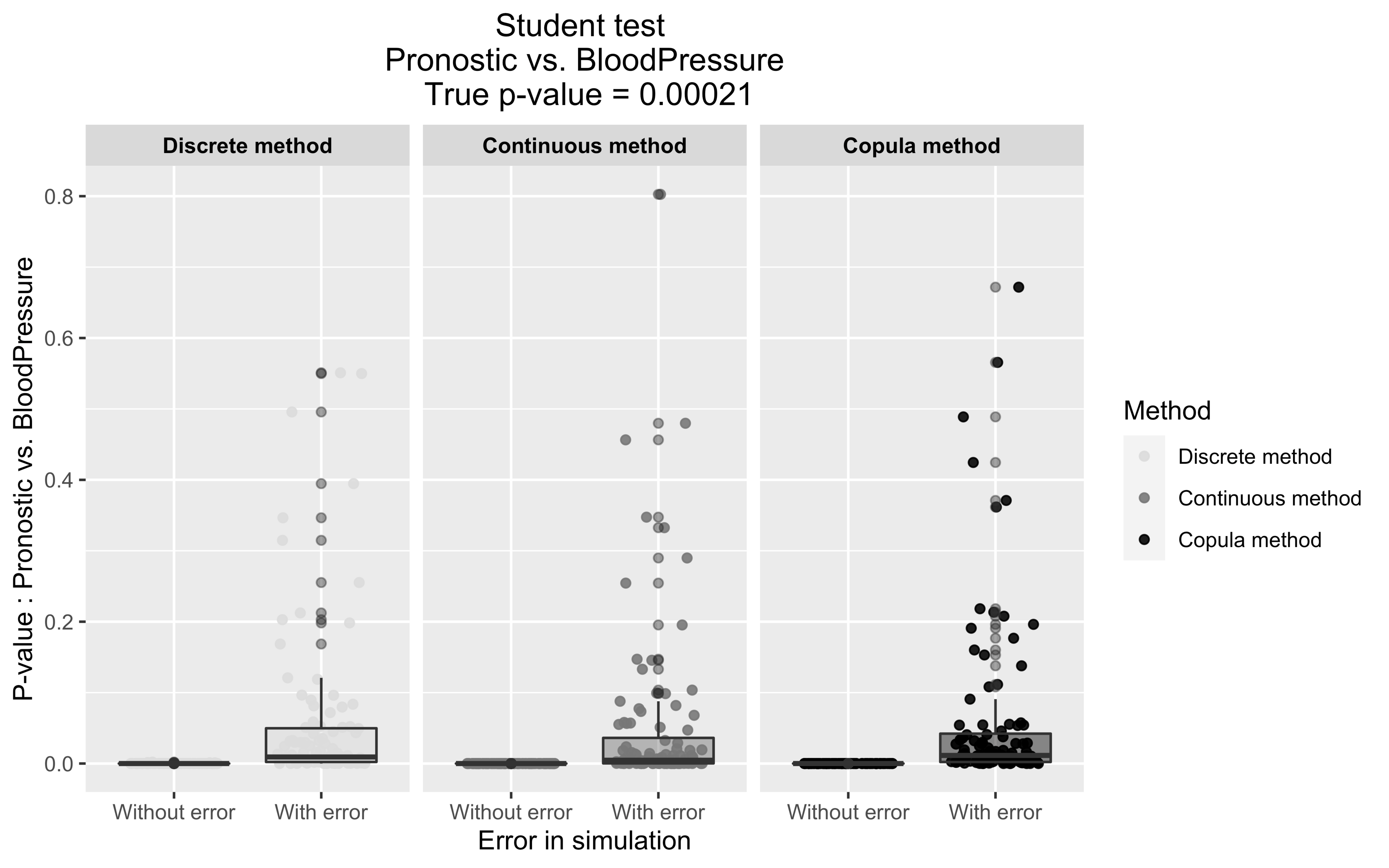}
		\caption{Simulation of $100$ datasets of $500$ observations each were performed using the discrete, the continuous and the Copula approaches. For each dataset, the p-value using Student test between \textit{Prognosis} variable and \textit{Blood pressure} were computed and reported using boxplot and jitter plot. The p-value observed in the original data is given in the title. Simulations are performed by adding or not an error in the prediction of the \textit{Prognostic} variable.}
		\label{fig:Pvalue_Var4}
\end{figure}

\begin{figure}
		\centering
		\includegraphics[width=1\textwidth]{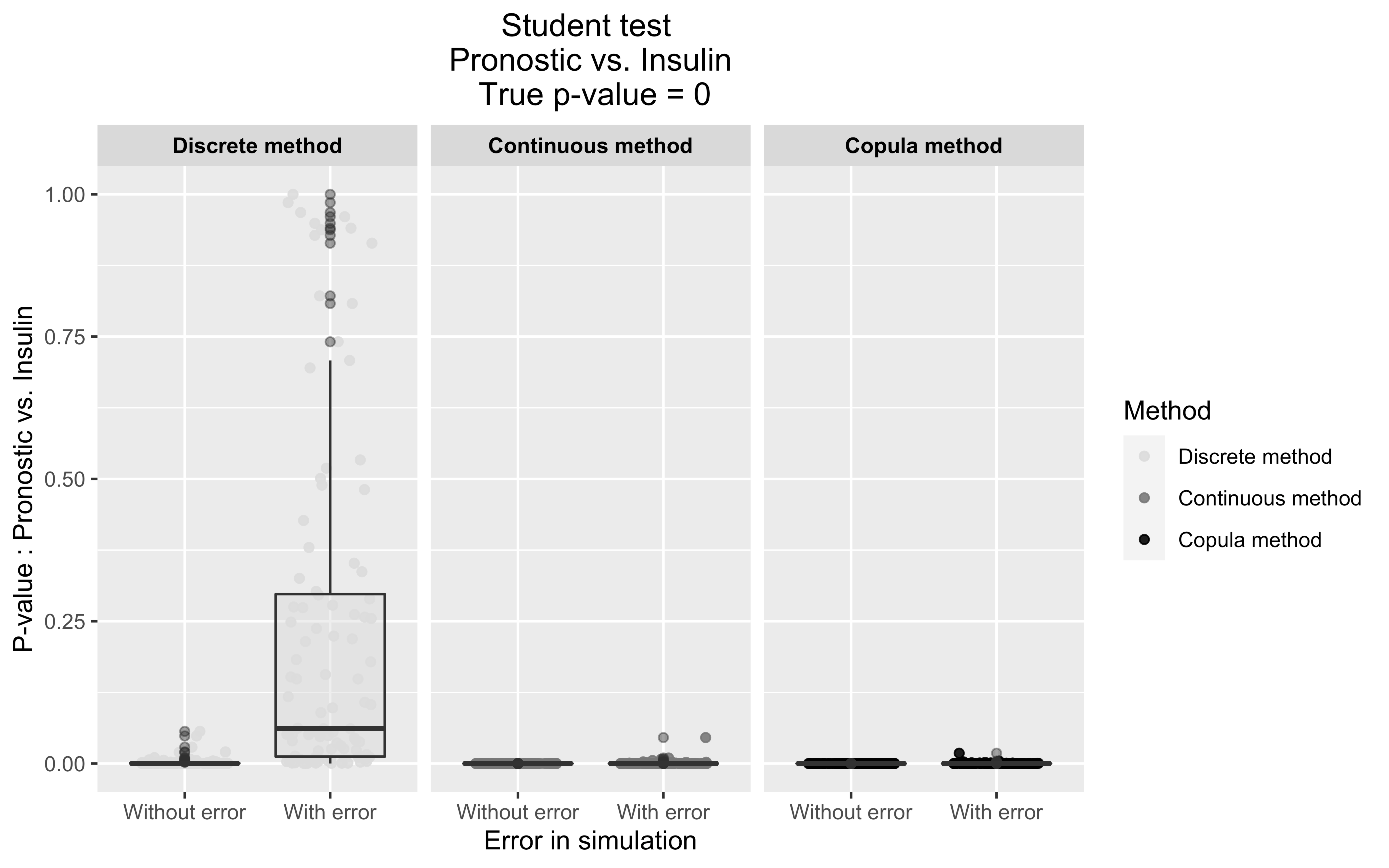}
		\caption{Simulation of $100$ datasets of $500$ observations each were performed using the discrete, the continuous and the Copula approaches. For each dataset, the p-value using Student test between \textit{Prognosis} variable and \textit{Insulin} were computed and reported using boxplot and jitter plot. The p-value observed in the original data is given in the title. }
		\label{fig:Pvalue_Var6}
\end{figure}

It is possible to evaluate the different simulation approaches by comparing the p-values of the association test between one variable and the prognosis variable in the original dataset and in the simulated dataset. For each simulation approaches, the p-values of the association test were evaluated for each dataset (Chi-2 test for categorical variables and Student test for quantitative variables). Figures \ref{fig:Pvalue_Var1}, \ref{fig:Pvalue_Var4} and \ref{fig:Pvalue_Var6} provide boxplot and jitter plot of the p-values according to the three simulation approaches for variables \textit{Pregnancy}, \textit{Blood pressure} and \textit{Insulin} respectively. Results are provided in two cases depending on whether prediction error (or noise) has been added to the simulated \textit{Prognostic} variable.

The Chi-2 p-value for \textit{Pregnancy} and \textit{Prognostic} in the original dataset is around 1 meaning an absence of relation (Figure \ref{fig:Pvalue_Var1}). Continuous and copula approaches provide relevant results whereas the discrete method is less coherent since the median of the p-value is the weakest and is significant when data are simulated without adding noise. One can observe a important variability of the p-values obtained for each simulation run for each approach. In some simulation runs, p-values are significant whereas there is clearly no relation in the original dataset. Is seems therefore important to simulate several dataset and to select  a nice one.

Figure \ref{fig:Pvalue_Var4} is a situation with a significant relation ($p=0.00021$) which is well identified and recovered by the three approaches. Even if the median of p-values is significant ($p<0.05$) with the three approaches, around $20\%$ of the simulation runs provide non significant p-values. 

Figure \ref{fig:Pvalue_Var6} present p-values in the case of a very strong association between \textit{Insulin} and \textit{Prognosis} in the original data. Continuous and Copula methods performed well since there are almost no p-values greater than $0.05$. However, in this case the discrete method fail to recover the association since the median of the p-values is not significant. 

This quick comparison of the three simulation approaches indicates that the discrete methods seems not satisfactory in this application certainly due to unbalanced situation. Despite the stringent Gaussian assumption, the results obtained with the continuous method are not so bad in this application maybe due to the uni-modality of the marginal distribution. Finally, Copula approach seems to provide the more relevant and promising results. Note that the differences observed between the original data and the simulated data can come from error in the simulation of the virtual patients but also from the error made when simulating the outcome (which is the same with the three methods).

\section{Example in Health Economic} \label{S-AEH}

\subsection{Context}

In France, the incidence of Human Immunodeficiency Virus (HIV) was estimated to be $6,600$ new cases per year and people living with HIV were estimated to be $156,600$ in 2014. Among them, 77\% were under AntiRetroViral treatment (ARV) \cite{S14}. As survival increased thanks to effective ARVs therapies, HIV became a chronic and costly disease. In the frame of the Long Term Disease (LTD) scheme, which allows reimbursement of a large part of disease-related costs, the French Health Insurance (FHI) estimated direct costs of HIV at 1.1 billion euros in 2009 \cite{T14}. Direct costs amounted to $13,000$ euros  per patient per year. ARV treatments accounted for 71\% of these costs \cite{CNAM}. Currently, the use of generic pharmaceutical products represents over half of the total volume of pharmaceutical products used worldwide, but only 18 \% of the pharmaceutical market total value \cite{S18}. In 2014 in France, the share of generics sold in the total medicines market was just over 30\% \cite{STA18}. From 2012 to 2016 only four generics of ARV were available. During the year 2017, three new generics of ARVs were launched on the market corresponding to combo generics which are more used than the first wave of generics. In the coming years, new generics of ARVs as well as patent expiry for other molecules will provide an opportunity for improving efficiency in the frame of constrained resource allocation.\\

Very few studies have assessed the economical impact of generic arrivals \cite{H14,R15}. These studies use, as most of the studies whom objective is the prediction of the cost in the context of a budget impact analysis, an approach based on an average behavior of patient and does not take into account inter-patients variability of the different parameters influencing the model. These population-based approaches yields to punctual predictions of the variations of costs embedded by the switch to the generics. In \cite{D21}, of which Section \ref{S-AEH} is a summary, authors have set up an Agent-Based Model to investigate the economical impact of of switching to generic in HIV medication. A patient is thus represented as a vector of covariates involving information on demography, medical status and treatments. Given these covariates, the behavior over time of each individual is simulated using execution models. Various models are involved mimicking the upadate of the covariates in time, the update of treatments, the status of the patient regarding comorbidities. These models are essentially Markov models. Each model makes use of two types of parameters: parameters fixed according to a predefined scenario and parameters of calibration estimated from the data (parameters of execution models). Given these individual simulated life trajectories, the treatment cost is evaluated under two scenarios: conversion to generics (according to a predefined scheme) and no conversion to generics allowing the evaluation of a differential cost. For the simulation to be relevant, it is necessary to set scenarios, depending on parameters which are fixed (not estimated from data), especially on the mode of conversion to the generics (marketing authorization date of the generics, penetration rate, tariffs,...).\\

This ABM is built from real-world observational data guaranting relevant outcome according to the French context we aimed to investigate. This approach allows to construct predictive intervals and to perform easily sensitivity analyses of outcomes of interest to parameters of the models. The full results are out of the scope of this paper which focuses on the methodological aspects. Complete results together with their health economical interpretation are discussed in \cite{D21}. However, as an illustration of the relevancy of this approach, results on the impact of penetration rate on the differential costs are presented in Section \ref{S-RES}.

\subsection{General statement of the Agent-Based Model}

\begin{figure}[h!]
\centering
\includegraphics[width=\textwidth]{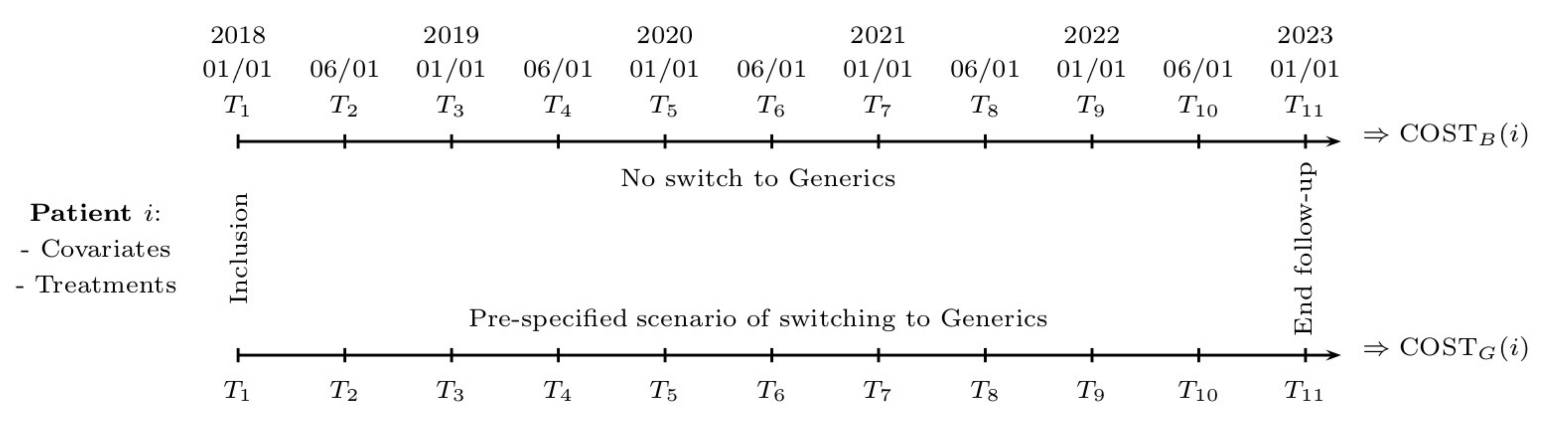}
\caption{Illustration of patient's follow-up.} \label{GEN}
\end{figure}

Let us consider cohorts of patients followed during 5 years by steps of six months which coincides with a standard medical monitoring of such patients. Follow-up times are denoted by $\left\{ T_u~: u=1, \dots, 11 \right\}$, an illustration of that timelines is given by Figure \ref{GEN}.\\

Data related to patient $i$ at times $T_u$ can be modeled as a vector $(\VPN{PATIENT}(u,i),$ $\VPN{TREAT}(u,i))$ of covariates where $\VPN{PATIENT}(u,i)$ stands for covariates associated to patient $i$ at time $T_u$ and $\VPN{TREAT}(u,i)$ stands for covariates associated to the treatments of patients $i$ at time $T_u$. This splitting between patients covariates and treatments covariates is only motivated by notational convenience. Notice that $(\VPN{PATIENT}(1,i),\VPN{TREAT}(1,i))$ consists in the baseline data for patient $i$ and play the role of initial values of the executions models involved to simulate patient $i$'s trajectories.\\

\begin{figure}[h!]
\centering
\includegraphics[width=0.75\textwidth]{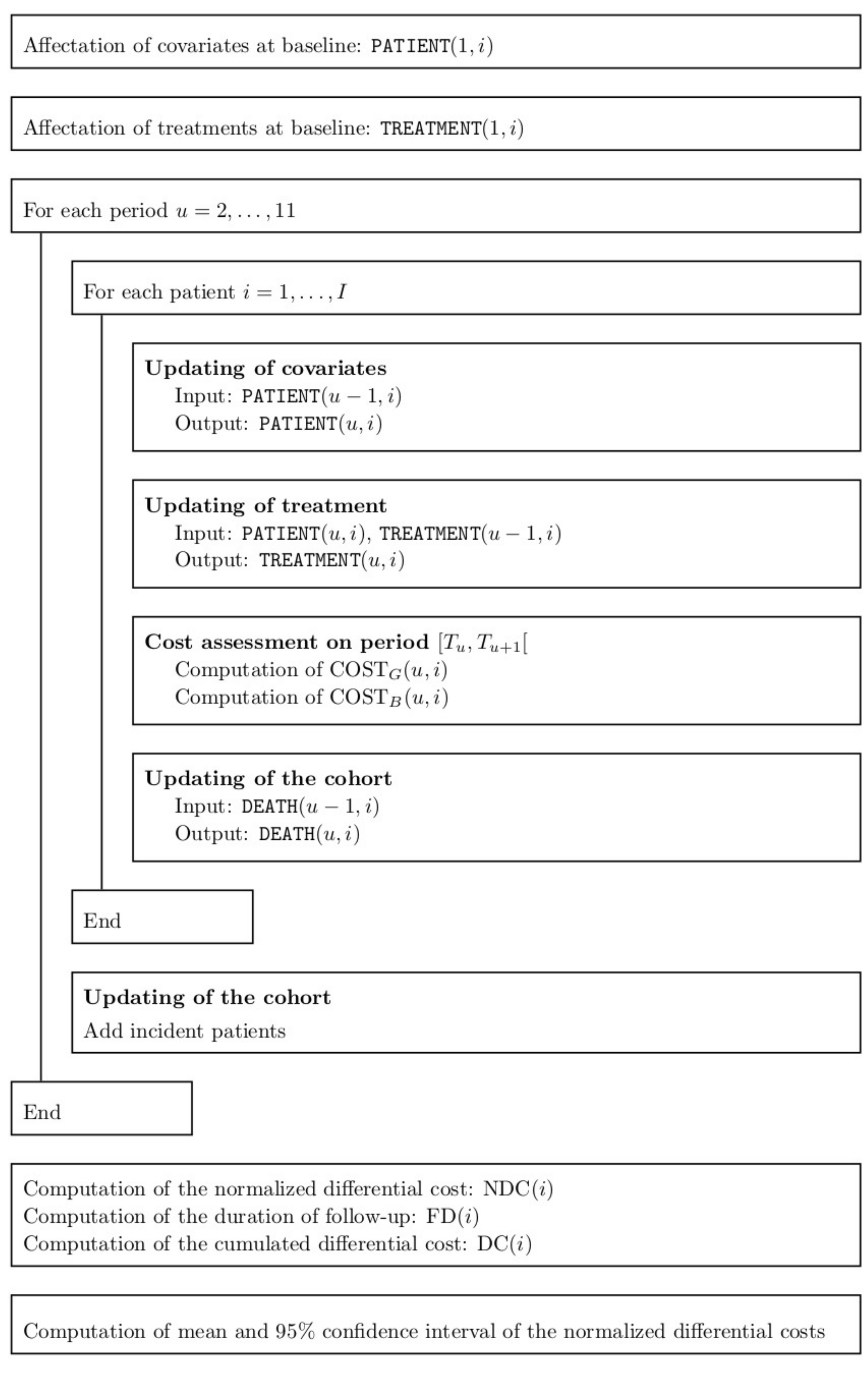}
\caption{Illustration of the algorithm.} \label{ALG}
\end{figure}

For each patient $i$, the patient trajectory is simulated thanks to execution models used to update its characteristics and its treatment. Patient $i$  follows two trajectories: a trajectory considering no switch to generics which generates a cost denoted $\OpN{COST}_B(i)$ and a trajectory assuming a pre-specified scenario of switching to generics $\OpN{COST}_G(i)$. In order to estimate the differential cost, for each time $T_u,$ $u \geq 2$,  the algorithm is split into four main steps:
\begin{itemize}
\item[Step 1:] Updating of the covariates of patient $i$:\\
 $\VPN{PATIENT}(u,i)$ is a modification of $\VPN{PATIENT}(u-1,i)$.
\item[Step 2:] Updating of the treatment of patient $i$:\\
 $\VPN{TREAT}(u,i)$ is a modification of $\VPN{TREAT}(u-1,i)$ according to $\VPN{PATIENT}(u,i)$, the  covariates of patient $i$ at time $T_u$,
\item[Step 3:] Updating of the status of the patient by considering possibility of death of patient $i$,
\item[Step 4:] Assessment of the costs for each scenario  during the period $[T_u,T_{u+1}[$, $\OpN{COST}_B(u,i)$ and $\OpN{COST}_G(u,i)$.
\end{itemize}
Differential costs and their prediction intervals can then be easily derived from the individual costs per period of time. A detailed illustration of the algorithm in given by Figure \ref{ALG}. Each step of the algorithm is modeled by an execution model and is specified in Section \ref{evolution}.

\subsection{Database used} \label{S-DATABASE}

In order to specify the characteristics of the patients of interest and to calibrate the executions models, several databases are used or constructed.\\

NADIS\textregistered~ (Fedialis Medica, Marly le Roi, France) is an Electronic Medical Record (EMR) for HIV-, hepatitis B virus (HBV)- or hepatitis C virus (HCV)-infected adults seeking care in French public hospitals \cite{PC2000}. NADIS\textregistered~ is currently used in $132$ french hospital centers. A technical staff and secretaries collect retrospective data during the first months of use of the EMR to ensure completeness of the patients records. Physicians prospectively provide information on patients records during each consultation or hospitalization. The resulting database, called $\BDN{NADIS}$ for a sake of simplicity, contains the following information for 27,341 patients.
\begin{itemize}
\item Civil status of the patient,
\item Social record,
\item HIV / Hepatitis clinical record,
\item Pathological and therapeutic history,
\item Clinical examinations,
\item Biological results,
\item Antiretroviral genotypes and dosings,
\item Medicinal prescriptions,
\item Examinations and checkup prescriptions,
\item Consultation and diagnosis motivations.
\end{itemize}

The $\BDN{MEDICATION}$ database is constructed from the $\BDN{NADIS}$ and involves 31 of the main medications used for HIV management. 
For each medication, the following information are collected:
\begin{itemize}
\item \texttt{NAMEM}: The name of the medication,
\item \texttt{REFO}: The amount refounded by Health Insurance,
\item \texttt{AMMT}: The marketing authorization date.
\end{itemize}
The database is also enriched with further information on the generic version of the treatment if it exists. This information is governed by parameters related to the simulation scenarios:
\begin{itemize}
\item \texttt{AMMGM}:  The marketing authorization date of the generic version of the medication. For HIV drugs, on average, the difference between the marketing authorization of the drug and its generic version is 13 years.
\item \texttt{PENRATE}: The maximal penetration rate of the generic version of the medication defined as the proportion of the population that consumes the generic  version of a medication. In 2012 the penetration rate of generics (all domains combined) reached 69.6\%  and the maximal rate is 80\% \cite{PC2000}.
\item \texttt{PENTIME}: The penetration time of the generic version of the medication defined as the time at which \texttt{PENRATE} is reached. The  probability of conversion to generic version (\texttt{PROBCONV}) is assumed to increase linearly between \texttt{AMMGM} and \texttt{PENTIME}. Figure \ref{PEN} page \pageref{PEN} illustrates the probability of conversion as a function of time.
\end{itemize}

The $\BDN{TREATMENT}$ database is constructed from the $\BDN{MEDICATION}$ database. In fact, a treatment is a combination of medications due to multi-therapy. This yields to a total of more than $800$ different treatments in the database. $\BDN{TREATMENT}$ database is composed of,
\begin{itemize}
\item \texttt{NAMET}: The name of the treatment which is a combination of medications.
\item \texttt{MEDCOSTB}: The baseline cost for the branded version of the treatment.
\item \texttt{MEDCOSTG}: The cost for the generic version of the treatment. The cost of the generic version of a drug is observed to be between 30\% and 50\% of the cost of the branded.
\end{itemize}
All drug prices are available on the VIDAL's website (https://www.vidal.fr). As the costs are time dependent, the tables are thus constructed with a column for each time step. To determine the price of a generic when it comes out, we know that it must be worth 40\% of the original tariff at this same date. This value is required by the government in the context of the current state-pharmaceutical industry framework agreement. It is also known that at the time the credits are released, the tariff of the branded drug decreases by 20\%. According to INSEE values, an annual decrease in tariffs of 3.4\% will also be taken into account, for both the branded and the generics drug.\\

It is important to point that a scenario of evolution will be parameterized by the choice of (\texttt{AMMGM}, \texttt{PENRATE}, \texttt{PENTIME}) and the values of the treatment costs (\texttt{MEDCOSTB} and \texttt{MEDCOSTG}).


\subsection{Baseline Cohort}

The  covariates involved in $\VPN{PATIENT}$ are the ones that may have an impact on the evolution of the disease, on the side-effects, on the comorbidities, on the evolution  of other covariates and on the choice of the medication. These covariates can be  classified in three categories:
\begin{itemize}
\item Demographic covariates,
\item Covariates linked to the pathology and its history,
\item Covariates linked to the comorbidities,
\end{itemize}
and are specified in Section \ref{S-COHORT}. Covariates linked to the treatment $\VPN{TREAT}$ integrate parameters which allow to investigate different scenarios and are specified in Section \ref{S-TREATVAR}.

\subsubsection{Description of patients covariates} \label{S-COHORT}

The cohort of patients consists in the whole patients followed in $\BDN{NADIS}$ (captured on December 31th 2015) from which are selected the covariates which may have an impact on the choice of the medication or on the evolution of other covariates. For each patient, the baseline values of these covariates may be directly captured in $\BDN{NADIS}$ or may be derived from covariates available in $\BDN{NADIS}$. These covariates can be  classified in three categories:\\

\textbf{Demographic covariates.} These covariates specify the main characteristics of the patients: $\VaN{SEX}$, the sex of the patient (0 for male and 1 for female), $\VaN{AGE}$, the age of the patient (in months) and $\VaN{BC}$, the country of birth, which differentiates patients born in France (modality $1$)  and patients born elsewhere (modality $0$). The values of $\VaN{SEX}(T_1)$, $\VaN{AGE}(T_1)$, $\VaN{BC}(T_1)$ are directly available in $\BDN{NADIS}$ database.\\

\textbf{Covariates linked to the pathology and its history.} $\VaN{CONTA}$, the way of contamination of the patient (1 for homosexual relationship and 0 for not), $\VaN{VIHS}$, the status of the infection (1 for SIDA and 0 for not). The duration of the HIV infection (in months),  denoted as $\VaN{VIHD}$, is a parameter of potential importance because during the first few months after the infection is discovered, the patient's treatment aims at reducing the number of copies of the virus in its organism. After a reduction of this number of copies below a certain threshold, the patient can receive a lighter treatment which aims at stabilizing this state of the infection. The duration of the last treatment, denoted as $\VaN{TREATD}$, (in months) is of major importance too. The longer is that duration, the lower is the probability to switch to another medication. Finally, the viral load, denoted as $\VaN{ARN}$, is an indicator of the progression of the disease and cannot be omitted in this study. The values of $\VaN{CONTA}(T_1)$, $\VaN{VIHS}(T_1)$, $\VaN{VIHD}(T_1)$, $\VaN{TREATD}(T_1)$ are directly available in $\BDN{NADIS}$ database. $\VaN{ARN}(T_1)$ is not available in $\BDN{NADIS}$ and is evaluated by an assessment of the viral load (a quantitative measurement denoted as $\VaN{ARNVIH}$ that exists in the database). This parameter is discretized in three modalities (0 for Low, 1 for Medium and 2 for High) according to thresholds recommended in \cite{ARN}:
$$
\text{$\VaN{ARN}(T_1)$} = 
\begin{cases}
0		&\text{if $0 \leq \VaN{ARNVIH}(T_1) < 50$}\\
1	 	&\text{if $50 \leq \VaN{ARNVIH}(T_1) < 10,000$}\\
2		&\text{if $\VaN{ARNVIH}(T_1) \geq 10,000$}
\end{cases}
$$

\textbf{Covariates linked to the comorbidities.} Cardiovascular illnesses, denoted as $\VaN{HEART}$  (1 for Yes and 0 for No), Diabetes, denoted as $\VaN{DIAB}$ (1 for Yes and 0 for No) and Renal failure, denoted as $\VaN{IR}$ (1 for Yes and 0 for No). The values of $\VaN{HEART}(T_1)$ and $\VaN{DIAB}(T_1)$ are directly available in $\BDN{NADIS}$ database. $\VaN{IR}$ is constructed as
$$
\text{$\VaN{IR}(T_1)$} = 
\begin{cases}
1			&\text{if $\VaN{CREA}(T_1)$ $ = 1$ or $\VaN{CREA}(T_1)$ $ = 2$}\\
0			&\text{if $\VaN{CREA}(T_1)$ $ = 3$}.
\end{cases}
$$
where $\VaN{CREA}$ (standing for Creatinine clearance) is a three modalities (0 for Low, 1 for Medium and 2 for High) categorical variable. $\VaN{CREA}$ is constructed thanks to the calculated glomerular filtration flow $\VaN{CGFF}$ that exists in the database:
$$
\text{$\VaN{CREA}(T_1)$} = 
\begin{cases}
1			&\text{if $\VaN{CGFF}(T_1)  > 89$}\\
2      		&\text{if $29 < \VaN{CGFF}(T_1)$ $ \leq  89$}\\
3			&\text{if $\VaN{CGFF}(T_1) \leq 29$}
\end{cases}
$$
The thresholds used are the ones recommended in \cite{CREA}.\\
The variable $\VaN{DEATH}$ indicates whether a patient is alive (1 for alive and 0 for dead) and is obviously initiated to 1. \\

Finally, vector of patient covariates at baseline $\VPN{PATIENT}(1,.)$ consists in the values at $T_1$ of the different covariates introduced below.

\subsubsection{Description of treatment variables} \label{S-TREATVAR}

The vector of patients treatment at baseline $\VPN{TREAT}(1,.)$ is directlty collected from $\BDN{NADIS}$ database and is enriched by the data related to the scenarios involved as described in Section \ref{S-DATABASE}.

\subsection{The execution models} \label{evolution}

Execution models are parts of the main algorithm in charge of mimicking the behavior of patients at each time step. Each baseline characteristic is then updated according to specific execution models to derive the values of $\VPN{PATIENT}(u,.)$ and $\VPN{TREAT}(u,.)$ for $u=2,...,10$. 

\subsubsection{Step 1: Updating of patients covariates} \label{S-COV}

To update the patients covariates at each time, different models are involved depending on the nature of the covariates and the precision that one aims at giving to the simulations:\\

\textbf{Covariates fixed in time.}
For $u=2, \dots, 10,$ we obviously have $\VaN{SEX}(T_u) = \VaN{SEX}(T_1),$ $\VaN{CONTA}(T_u) = \VaN{CONTA}(T_1),$ $\VaN{BC}(T_u)= \VaN{BC}(T_1)$.\\

\textbf{Covariates with deterministic dependence on time.} $\VaN{AGE}$ and $\VaN{VIHD}$ change during the follow-up of patient because time is going on and these covariates are nothing but duration. The changes are nothing more than an increment of six months in the covariates. For $u=2, \dots, 10,$ we have $\VaN{AGE}(T_u) = \VaN{AGE}(T_{u-1}) + 6,$ and $\VaN{VIHD}(T_u) = \VaN{VIHD}(T_{u-1}) + 6$. $\VaN{TREATD}$ evolves in the same way but have to be reset to 0 in case the treatment is changed. For $u=2, \dots, 10,$ we have:
$$
\text{$\VaN{TREATD}(T_u)$} =
\begin{cases}
 \VaN{TREATD}(T_{u-1}) + 6, \quad \text{if there is no switch of treatment,}\\
0, \quad \text{if there is a change of treatment at time $T_u$}.
\end{cases}
$$

\textbf{Covariates with random dependence in time.} Covariates $\VaN{HEART}$, $\VaN{DIAB}$, $\VaN{VIHS}$ and $\VaN{DEATH}$ may change during the patients follow-up. These changes can lead to a modification of patient treatment. The evolution of these covariates are directed by Markov chains where the matrices of transition, denoted $M_{\VaN{HEART}}, M_{\VaN{DIAB}}$, $M_{\VaN{VIHS}}$ and $M_{\VaN{DEATH}}$ are chosen to be constant.\\

\textbf{Covariates with random dependence in time and randomness depending of covariates.} 
For $\VaN{ARN}$ and  $\VaN{CREA}$, the matrices of transition, denoted  $M_{\VaN{ARN}}$ and  $M_{\VaN{CREA}}$ cannot be assumed to be constant because these transitions depends on patients covariates. For these models, the probabilities of transition are modeled by a logistic or a polytomic regression. The covariates involved in the model are selected by a backward stepwise strategy. For $\VaN{ARN}(T_u)$, those are $\VaN{ARN}(T_{u-1})$, $\VaN{IR}(T_{u-1})$, $\VaN{CONTA}(T_{u-1})$, $\VaN{HEART}(T_{u-1})$, $\VaN{VIHS}(T_{u-1})$, $\VaN{AGE}(T_{u-1})$, $\VaN{SEX}(T_{u-1})$, $\VaN{VIHD}(T_{u-1})$ and $\VaN{TREATD}(T_{u-1})$ and for $\VaN{CREA}(T_u)$ the covariates are $\VaN{CREA}(T_{u-1})$, $\VaN{SEX}(T_{u-1})$, $\VaN{ARN}(T_{u})$, $\VaN{AGE}(T_{u-1})$, $\VaN{HEART}(T_{u-1})$, $\VaN{TREATD}(T_{u-1})$, $\VaN{VIHS}(T_{u-1})$ and $\VaN{VIHD}(T_{u-1})$.\\

\textbf{Calibration of the execution models.} The calibration of the models which means the estimation of the coefficients of $M_{\VaN{HEART}}, M_{\VaN{DIAB}}$, $M_{\VaN{VIHS}}$ and  $M_{\VaN{DEATH}}$ as well as the estimation of the parameters of the logistic (polytomic) regressions involved in the coefficients of  $M_{\VaN{ARN}}$ and  $M_{\VaN{CREA}}$ are derived from $\BDN{NADIS}$ database.\\

A summary of the covariates considered in this simulation plan and their related execution models is presented in Table~\ref{COV}.

\begin{table}
\caption{List of covariates involved together with the associated modalities and execution models. LIN refers to LINear model, C to Constructed model, LBP to Linear By Part model, MC to Markov Chain and MCRT to Markov Chain with Random Transitions.} \label{COV}
\centering
\begin{tabular}{l c c c c}
\hline
Covariate &Name &Modalities &Execution\\
\hline
Age 															&$\VaN{AGE}$			&N/A 										&LIN \\
Cardiovascular illnesses						&$\VaN{HEART}$ 	&YES / NO							&MC \\
Country of birth (France)					&$\VaN{BC}$   			&YES / NO							&N/A \\
Creatinine clearance							&$\VaN{CREA}$ 		&Low / Medium / High		&MCRT  \\
Death         												&$\VaN{DEATH}$ 	&YES / NO							&MC \\
Diabetes													&$\VaN{DIAB}$ 		&YES / NO							&MC \\
Duration of the last treatment			&$\VaN{TREATD}$ 	&N/A 	 									&LBP \\
Duration of the VIH infection 			&$\VaN{VIHD}$		    &N/A 		 								&LIN \\
Renal failure											&$\VaN{IR}$ 				&YES / NO							&C \\
Sex 																&$\VaN{SEX}$  		    &Male / Female					&N/A \\
Status of VIH infection (SIDA)			&$\VaN{VIHS}$  		&YES / NO							&MC \\
Viral load													&$\VaN{ARN}$  		&Low / Medium / High 	&MCRT \\
Way of contamination (Hom. Rel.)	&$\VaN{CONTA}$  	&YES / NO							&N/A \\
\hline
\end{tabular}
\end{table}

\subsubsection{Step 2: Updating the treatment}  \label{S-TREAT}

The updating of the treatment at time $T_u$, that is formalized by column $\VPN{TREAT}(u,.)$, follows four rules:
\begin{itemize}
\item[Rule 1:] a patient can keep his treatment,
\item[Rule 2:] a patient can switch for a treatment to another,
\item[Rule 3:] a patient can switch to the generic version of his treatment,
\item[Rule 4:] a patient who converts its medication to generic cannot convert back to the branded as long as he does not change its treatment.
\end{itemize}

\begin{figure}[h!]
\centering
\includegraphics[width=0.9\textwidth]{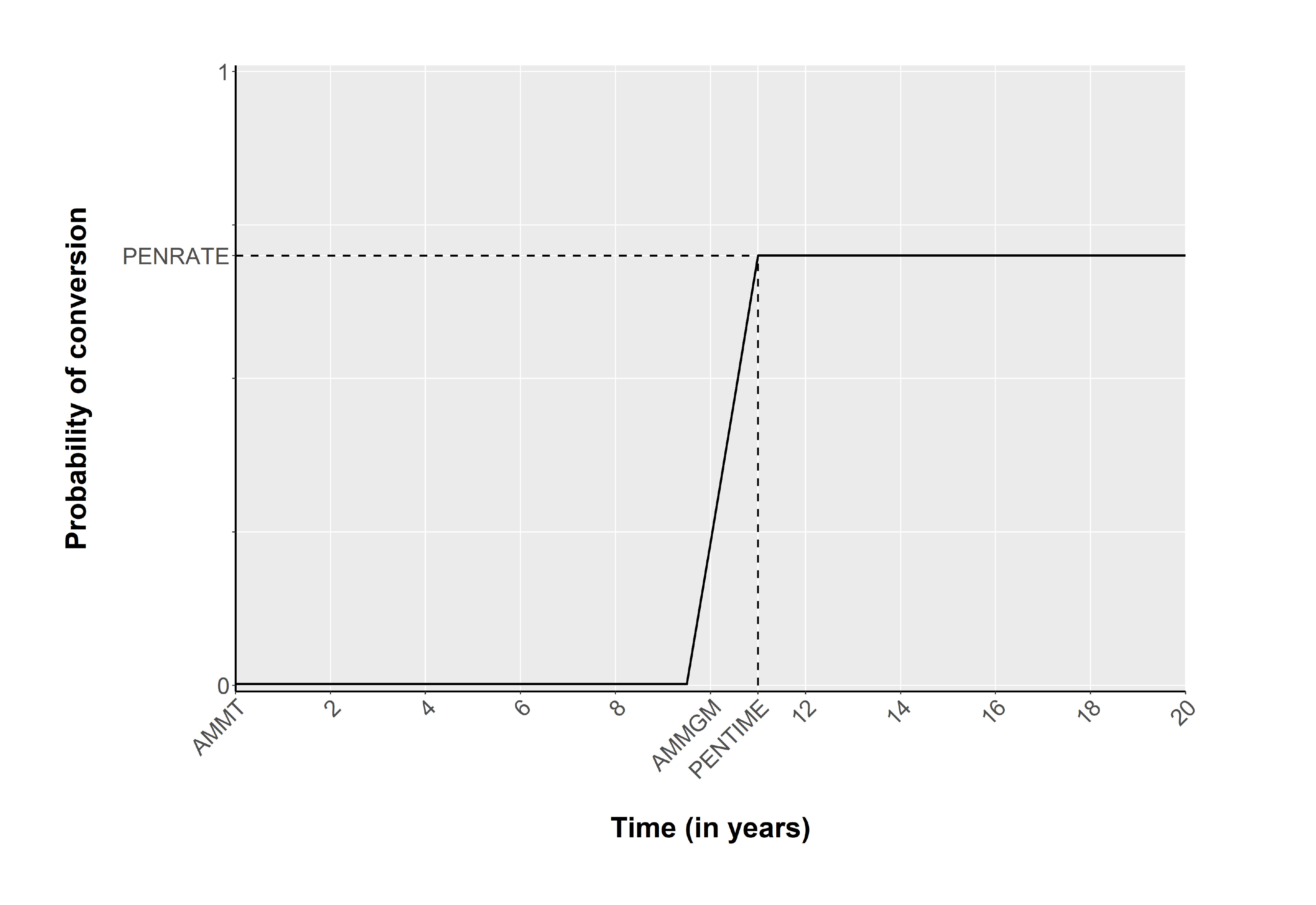}
\caption{Evolution of the probability of conversion through time with a marketing authorization date to date of generic fixed at 10 years after the marketing authorization date of the branded drug.} \label{PEN}
\end{figure}

The execution models, accounting for these rules, are defined as:
\begin{itemize}
\item Patients can switch for a treatment to another according to a Markov chain where the transition matrix is estimated from the $\BDN{NADIS}$ database. For those transitions observed for a large number of patient in the $\BDN{NADIS}$, a logistic regression model is adjusted in order to model the transition probabilities as a function of some covariates. In this setting, the choice of covariates involved is done case-by-case following a backward step-by-step strategy. When transitions are rare (less then $100$ observations), their probability are considered as constant.
\item Patients can switch to the generic version of his treatment. This happens with probability depending of time $t$ represented by Figure \ref{PEN} and defined by:
$$
\text{\texttt{PROBCONV}(t)} = 
\begin{cases}
0 														&\text{if $t < $\texttt{AMMGM}},\\
\text{\texttt{PENRATE}}	\frac{t - \text{\texttt{AMMGM}}}{\text{\texttt{PENTIME}} - \text{\texttt{AMMGM}} } 							&\text{if \texttt{AMMGM}$\leq t <$ \texttt{PENTIME}},\\
\text{\texttt{PENRATE}}			&\text{if $ t > $\texttt{PENTIME}}.
\end{cases}
$$
\end{itemize}

\subsubsection{Step 3: Updating of the cohort}

On the one hand, at a given time step $T_u$, an updating of the cohort is performed by including 455 incident cases. This quantity is the ones found in $\BDN{NADIS}$ database and is in adequacy with literacy results \cite{LOT}. These incident cases are randomly chosen from the $\VPN{PATIENT}(1,.)$ vector. The treatment for each incident case is randomly chosen in the updated treatment variable at timue $u$, $\VPN{TREAT}(u,.)$ in order to account for the evolution of prescriptions and prices in time.\\
 
On the other hand, a patient $i$ who died during the period $[T_{u-1}, T_{u}[$ this means a patient for which $\VaN{DEATH}(u-1,i) = 1$ and $\VaN{DEATH}(u,i) = 0$, is not removed but his future costs are fixed to 0.\\

Finally, let us denote $\OpN{FD}(i)$ the duration of the follow-up (in semester) for patient $i$.

\subsubsection{Step 4 : Computation of the differential cost}

The differential cost $\OpN{DC}$ for patient $i$ is defined by:
$$
\OpN{DC}(i) = \OpN{COST}_B(i) - \OpN{COST}_G(i) = \sum_{u=1}^{10} \left(  \OpN{COST}_B(u,i) - \OpN{COST}_G(u,i) \right),
$$
where $\OpN{COST}_B(i)$ represents the cost for patient $i$ considering no switch to generics and $\OpN{COST}_G(i)$ represent the cost assuming a pre-specified scenario of switching to generics. The main indicator of interest is the total differential cost on five years for French population defined as the sum over patients of the individual differential cost $\OpN{DC}(i)$ divided by $22.8\%$ which is an approximation of the fraction of the french HIV population integrated in $\BDN{NADIS}$. Notice that if a patient $i$ died during the period $[T_{u}, T_{u+1}[$ his future costs are fixed to 0, this means $\OpN{COST}_B(v,i) = \OpN{COST}_G(v,i) = 0$ for any $v \geq u+1$.\\

As described later patients may die before the end of the follow-up and incident cases are possible. The differential cost should be normalized according to the follow-up duration to define the normalized differential cost:
$$
\OpN{NDC}(i) = \frac{\OpN{DC}(i)}{\OpN{FD}(i)}.
$$
Another indicator of interest is the average differential cost per patient per semester and defined as the average over the patients of the values of $\OpN{NDC}(i)$.

\subsection{Example of results} \label{S-RES}

To illustrate the relevancy of this approach, results on the impact of penetration rate on the differential cost are presented. This is a result of major importance in this context. The full results  are out of the scope of this paper which focuses on the methodological aspects of the agent-based simulation approach especially a statistical point of view focusing on the variability issue. Completed results together with their health economical interpretation are presented in \cite{D21}. These results include, for example, sensitivity analyses on the marketing authorization date, on the decrease in tariffs.

\subsubsection{Scenarios investigated.}

The scenarios investigated are the following ones, for each treatment $k$:
\begin{itemize}
\item \texttt{AMMGM(k):} is fixed to 13 year after \texttt{AMMT}, whatever $k$,
\item \texttt{PENTIME(k):} is fixed to one year after $\texttt{AMMGM}$, whatever $k$,
\item \texttt{PENRATE(k):} five penetration rates (10\%, 25\%, 40\%, 55\% and 70\%) are considered whatever $k$.
\item \texttt{MEDCOSTB}: the cost at baseline is the cost refunded by Health Insure the first semester 2018.
\item \texttt{MEDCOSTG}: is fixed to 40\% of the branded cost at \texttt{AMMGM}, whatever $k$.
\end{itemize}
In order to get an idea of the relevance of the predictions, $100$ simulation runs are performed (note that an higher number of simulation runs does not change the results) yielding to the empirical distribution of each parameter from which it is easy to derive 90\% (resp. 80\%) prediction intervals considering the 5th and 95th (resp. 10th and 90th) values of the sorted distribution.

\subsubsection{Results obtained} 

The results obtained are the following. It is important to notice that such results cannot be obtained without an agent-based modeling.
\begin{itemize}
\item The estimate of the total differential cost on five year for French population together with their 90\% and 80\% prediction intervals are illustrated by means of boxplots in Figure \ref{F-ILLU}.
\begin{figure}[h!]
\centering
\includegraphics[width=0.9\textwidth]{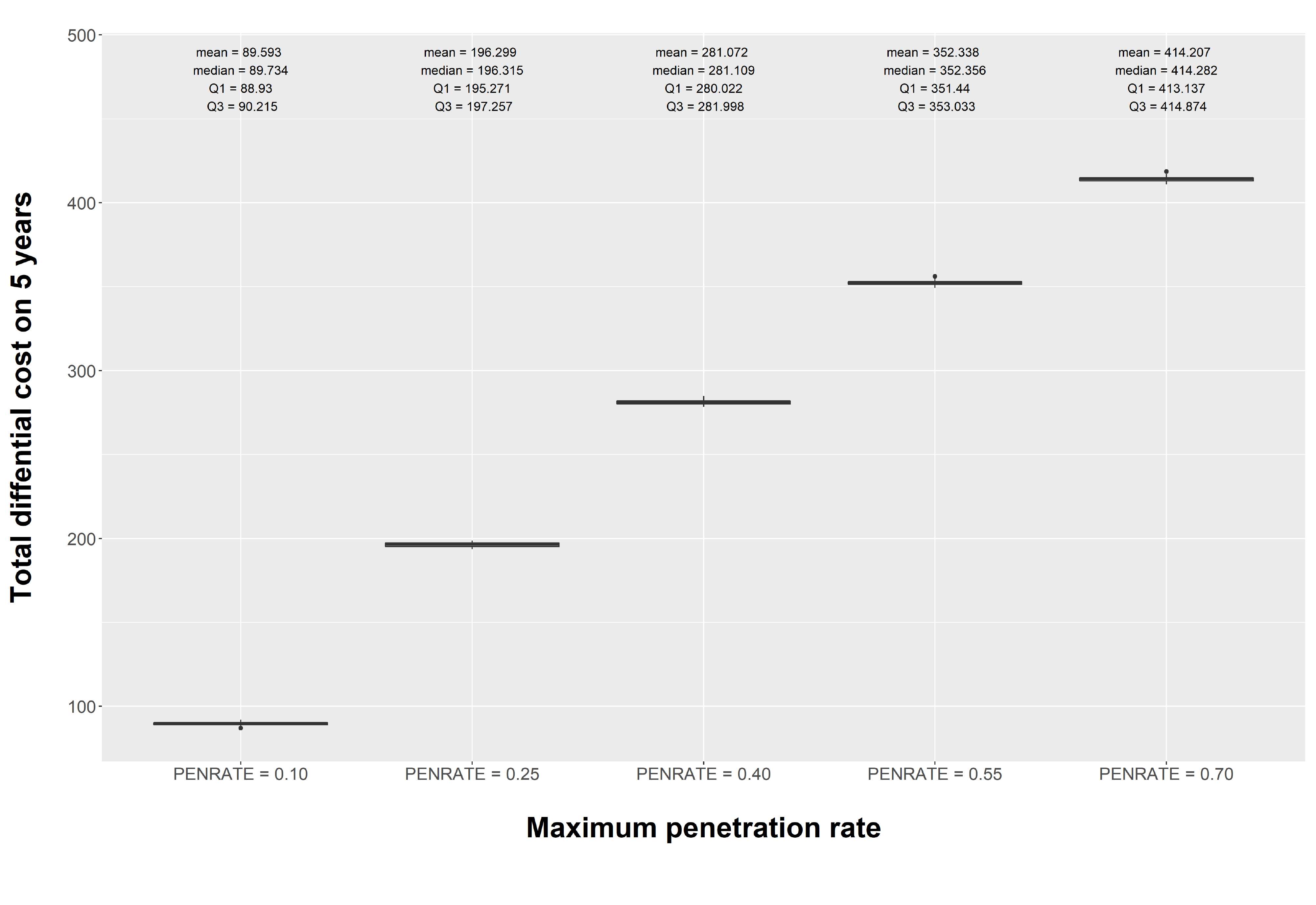}
\caption{Boxplot of the total differential cost on five year for French population as a function of the penetration rate (from $100$ simulation runs, in millions on euros).}
\label{F-ILLU}
\end{figure}
\item The estimate of the normalized differential cost per patient per year together with their 90\% and 80\% prediction intervals are illustrated by means of boxplots in Figure \ref{F-ILLU2}.\\
\begin{figure}[h!]
\centering
\includegraphics[width=0.9\textwidth]{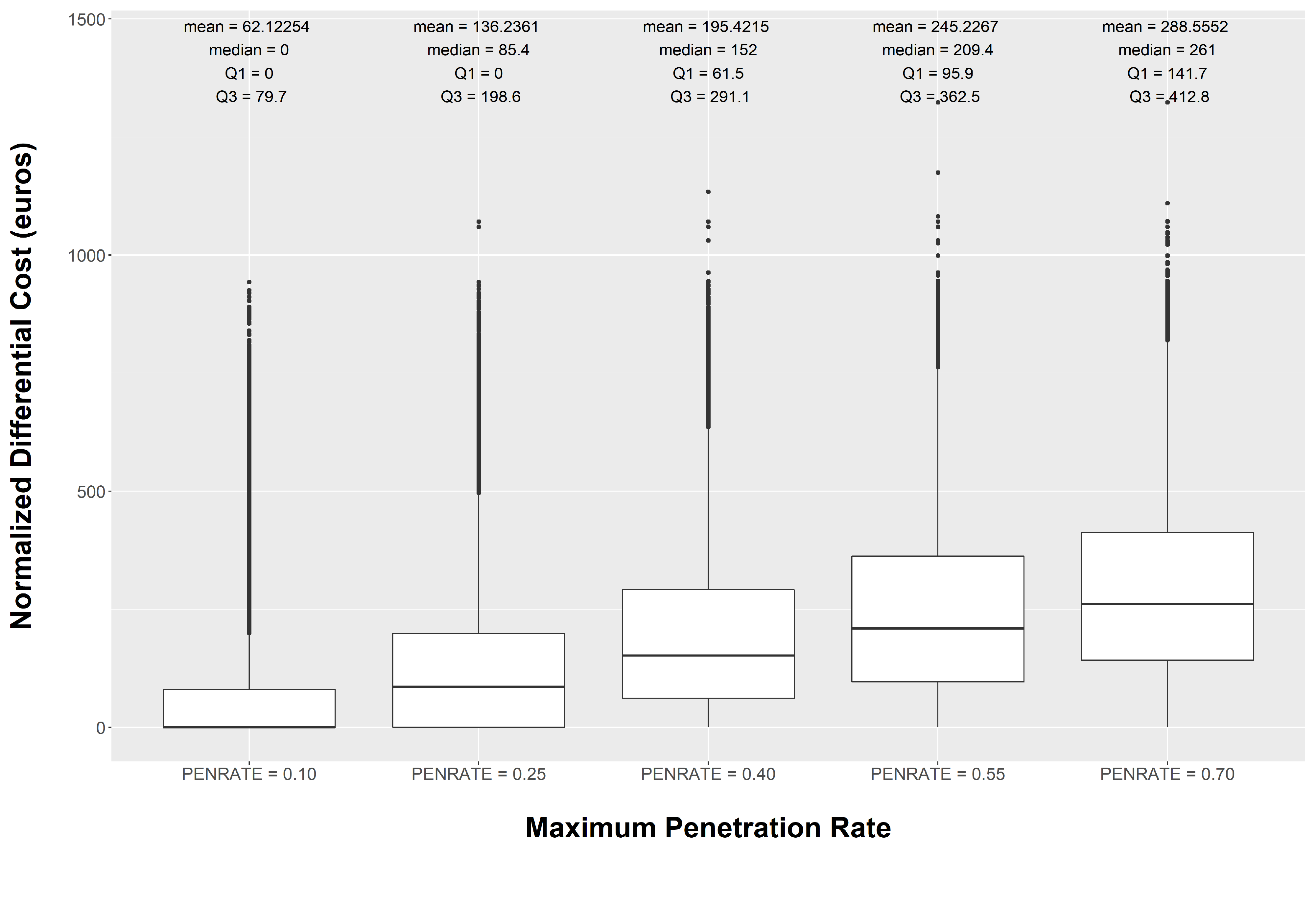}
\caption{Boxplot of the normalized differential costs per patient per year as a function of scenarios defined by penetration rate (from $100$ simulation runs, in euros).}
\label{F-ILLU2}
\end{figure}
These results are enriched by  Figure \ref{F-MCD} which represented, for each scenario, the distributions of the normalized differential cost per patient obtained after $100$ simulation runs. These plots give us a preview of the distribution of that value in the population for a given simulation run.
\begin{figure}[h!]
\centering
\includegraphics[width=0.9\textwidth]{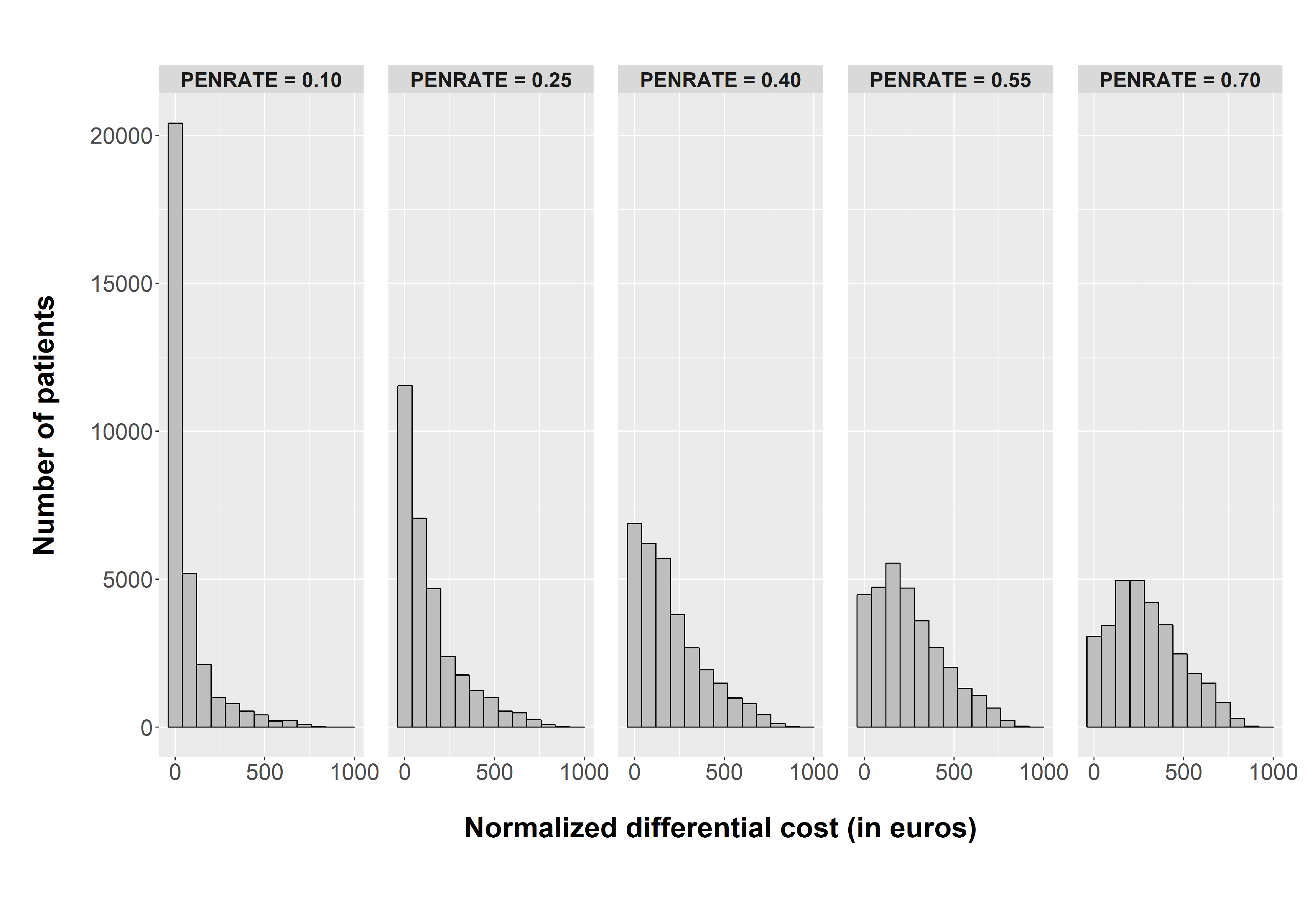}
\caption{Distribution of the normalized differential costs per patient as a function of scenarios defined by penetration rates (from $100$ simulation runs, in euros).}
\label{F-MCD}
\end{figure}
\item The estimate of the proportion of patients who were prescribed to generic at least once during the follow-up together with their 80\% prediction intervals are illustrated by means of boxplots in Figure \ref{F-ILLU3}.
\begin{figure}[h!]
\centering
\includegraphics[width=0.85\textwidth]{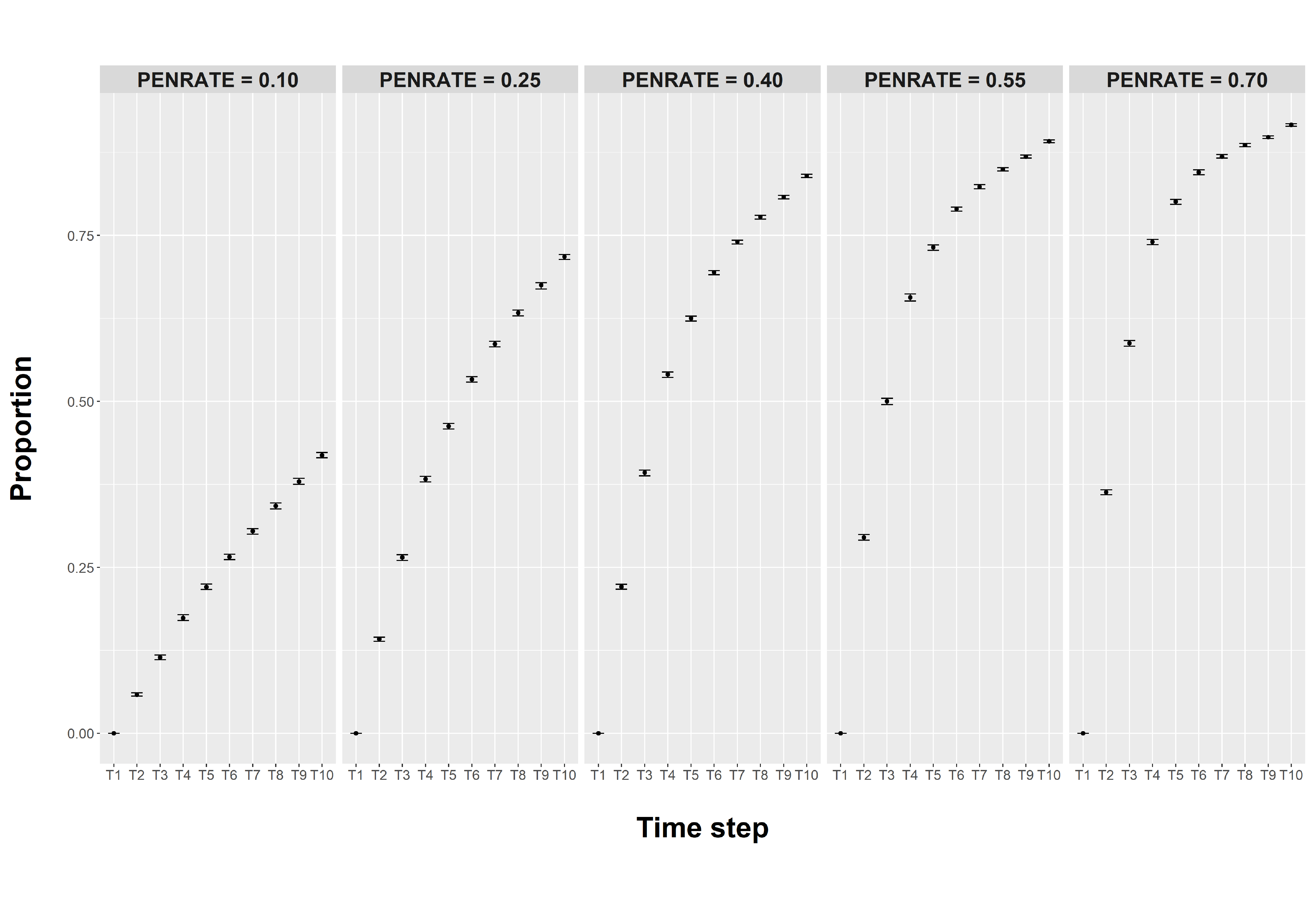}
\caption{Boxplots of the proportion of patients who were prescribed a generic at least once during the follow up together with their prediction intervals (from $100$ simulation runs). 80\% prediction intervals.}
\label{F-ILLU3}
\end{figure}
\end{itemize}

First of all, the differential costs naturally increases with the penetration rate but this increasing is not linear. Notice that the precision in the prediction assessed by the standard deviation of predicted values increases very slowly with the penetration rate. The main result here is the comparison between the results on the total differential costs which is a parameter related to population (Figure \ref{F-ILLU}) and the per patient differential cost (Figure \ref{F-ILLU2}). Results for total differential cost yields to very small prediction interval whereas results for a given patient (Figure \ref{F-MCD}), yields to very large prediction intervals. That is not so surprising because the results for total differential cost aggregate the effect of patients of whole the cohort. This comparison is of major importance because it highlights the fact that even if the prediction is quite good at the population scale, this covers a huge variability at the individual scale. Roughly speaking, the differential cost  begins to be noticeable only for a large penetration rate. Such results cannot be observed by means of a population-based approach. Exactly as for total differential costs, it is not surprising to observe very small prediction intervals (Figure \ref{F-ILLU3}). Notice that there is no difference between 90\% and 80\% confidence intervals, the prediction are very narrow. Finally, it is of paramount interest to observe those patterns to see how the proportion of patients who switch to generic is modified by the maximal penetration rate. We can also notice that 0 is contained in all prediction intervals, except at time T10 for a penetration rate of 70\%.  This observation is highly correlated with the proportions of patients who switch to generics. In fact, as can be seen in Figure \ref{F-ILLU3}, the only proportion that exceeds 90\% is the one obtained at time T10 with a penetration rate of 70\%.

\subsection{Discussion and conclusions} \label{S-DIS}

The agent-based method proposed here in the context of health economics research presents four major advantages:
\begin{itemize}
\item First, it allows to integrate, in the prediction of the differential costs, much more parameters especially individual parameters together with their correlation structures making the predictions more realistic.
\item Second, it allows to study the effect of time by considering longitudinal models.
\item Third, it allows to assess the precision of the predictions by incorporating randomness in the dynamic of the system. That precision can be evaluated or illustrated by means of prediction intervals coming from the distribution of the predictions obtained through several simulation runs. It is a precious tool in this context to compare and to identify the main sources of randomness.
\item Fourth, it allows to investigate individual behavior of patients and thus to modulate the conclusion in terms of population conclusions together with individual conclusions. The randomness of the proposed dynamic system comes from two points:
\begin{itemize}
\item the several simulation runs to observe the variability intra-patient with potentially very different behavior from a run to another,
\item the dynamic in time (simulation of patient's trajectory via execution models) for a given simulation run.
\end{itemize}
\end{itemize}
The first point is not predominant when dealing with total cost because whole the population is involved in the prediction by aggregation.\\

The main drawback is usual in modeling and comes from the  choice of the models involved and the performances of the calibration of such models. The choice of the models is driven by the data and by the clinical input on the disease. For this study, we strongly benefited from the help of $\BDN{NADIS}$ scientific committee, composed of experts in the management of HIV-infected patients.

The results presented here are conditional to assumptions underlying the execution models which are detailed in the associated section. Notice that we did not take into account the ability of some patients to break their combos (\textit{i.e.} switching from a one pill combination of several medications to several pills) as it would have resulted in a much higher algorithm complexity. Such a consideration ensures that when the generic version of a medication that is also part of a combo is available, but the combo itself is not, patients are prevented from breaking it to take the generic. Hence, resulting cost savings may be underestimated.\\

It is a commonplace to say that agent-based approach is a balance between complexity of the model (which insures its relevancy) and its use. Indeed the more a model is complex, the more difficult to calibrate it is and the more sensitive to the model the results are. Note that the execution models can easily be adapted and complexified to take into account the specificities of the disease and of the treatments. For example, it is possible to consider time-dependent covariate logistic regressions or time-dependent covariate Markov models. Such models could then be adapted and implemented in other chronic diseases in order to simulate potential economic savings due to switching to generics or to any modification in treatments costs. These savings could be used to fund prevention care or innovative care which provides better health to the population in terms of quantity and quality.

\section{Conclusion, Recommendations and Take home message} \label{S-RECO}

The main elements of an ABM are the virtual baseline generator and the execution models. Both elements have their own constraints: the virtual baseline generator must simulate data with realistic correlation structure and marginal distributions and the execution models must simulate virtual outcomes accounting for the error of prediction.

Thinking about data is of paramount importance in building an ABM. Several elements have to be taken into account: the choice of variables, the availability of data, the volume of data or the origin of the data (real word, clinical trials, cohorts, etc.). In practice, this information about the data indicates the execution models that can be implemented (e.g. minimal volume is required to implement most of machine learning algorithms).

One must paid particular attention to the chronology in the sequence of the execution models. Indeed, in practice, the sequence of execution models  may have a significant impact on the result of the simulations.

Sources of variability in ABMs come from the generation of the baseline data (if any) and the prediction error of the execution models. These errors are cumulative and can lead to very noisy situations. To ensure consistent results, it is necessary to run the simulations several times and provide the results in terms of parameters constructed from the empirical distribution generated by the different runs (median, prediction interval, boxplot...).

Building ABMs necessitate a strong interdisciplinary approach  including statisticians (for building models), health data specialists (for identifying and locating databases) and clinicians (for identifying the variables of interest and defining the execution models to consider).

One can note that ABMs  presented here can be extended to multi-agent models, especially in an epidemic context where the agents are not independent. The situation is more complex in terms of implementation. There exists dedicated tools to design an ABM as \textit{NetLogo} software which can then be integrating in R software via the \textit{RNetLogo} package \cite{RNtetLogo}.

\newpage

\section*{Acknowledgements}

Authors warmly thank the Study Group Dat'Aids:\\

\textbf{Besan\c con :} C. Drobacheff-Thi\'ebaut, A. Foltzer, K. Bouiller, L. Hustache- Mathieu, C. Chirouze, Q. Lepiller, F. Bozon, O Babre, A.S. Brunel, P. Muret. \textbf{Clermont-Ferrand :} H. Laurichesse, O. Lesens, M. Vidal, N. Mrozek, C. Aumeran, O. Baud, V. Corbin, P. Letertre-Gibert, S. Casanova, J. Prouteau, C. Jacomet. \textbf{Guadeloupe :} I. Lamaury, I. Fabre, E. Curlier, R. Ouissa, C. Herrmann-Storck, ,B. Tressieres, T. Bonijoly, M.C. Receveur, F. Boulard, C.Daniel
% ...
M. Valette.

\end{document}